\documentclass[journal]{new-aiaa}
\usepackage[utf8]{inputenc}
\usepackage{textcomp}

\usepackage{graphicx}
\usepackage{amsmath}
\usepackage[version=4]{mhchem}
\usepackage{siunitx}
\usepackage{longtable,tabularx}
\usepackage{booktabs}
\usepackage{cleveref}
\usepackage{subcaption}
\usepackage{algorithm}
\usepackage[noend]{algpseudocode}
\usepackage{tikz}
\usetikzlibrary{shapes,arrows,positioning,hobby,backgrounds,calc,trees}

\pgfdeclarelayer{background}
\pgfsetlayers{background,main}

\hypersetup{hidelinks}

\NewDocumentCommand{\angsi}{omom}{%
    \ang[#1]{#2}\,\si[#3]{#4}%
}

\usepackage{color}

\title{A Maximum Independent Set Method for Scheduling Earth Observing Satellite Constellations}

\author{Duncan Eddy\footnote{PhD Candidate, Aeronautics and Astronautics, Durand Building 450 Lomita Mall, and AIAA Student Member.} and Mykel J. Kochenderfer\footnote{Associate Professor, Aeronautics and Astronautics, Durand Building 450 Lomita Mall, and AIAA Associate Fellow.}}
\affil{Stanford University, Stanford, CA, 94305, USA}

\begin{document}

\maketitle

\begin{abstract}

Operating Earth observing satellites requires efficient planning methods that coordinate activities of multiple spacecraft.  The satellite task planning problem entails selecting actions that best satisfy mission objectives for autonomous execution. Task scheduling is often performed by human operators assisted by heuristic or rule-based planning tools. This approach does not efficiently scale to multiple assets as heuristics frequently fail to properly coordinate actions of multiple vehicles over long horizons. Additionally, the problem becomes more difficult to solve for large constellations as the complexity of the problem scales exponentially in the number of requested observations and linearly in the number of spacecraft. It is expected that new commercial optical and radar imaging constellations will require automated planning methods to meet stated responsiveness and throughput objectives. This paper introduces a new approach for solving the satellite scheduling problem by generating an infeasibility-based graph representation of the problem and finding a maximal independent set of vertices for the graph. The approach is tested on a scenarios of up to 10,000 requested imaging locations for the Skysat constellation of optical satellites as well as simulated constellations of up to 24 satellites. Performance is compared with contemporary graph-traversal and mixed-integer linear programming approaches. Empirical results demonstrate improvements in both the solution time along with the number of scheduled collections beyond baseline methods. For large problems, the maximum independent set approach is able find a feasible schedule with 8\% more collections in 75\% less time.
 
\end{abstract}

\section*{Nomenclature}

{\renewcommand\arraystretch{1.0}
\noindent\begin{longtable*}{@{}l @{\quad=\quad} l@{}}
$S$ & set of all constellation spacecraft \\
$s$ & single spacecraft in constellation \\
$H$ & planning horizon duration \\
$R$ & set of all tasking (observation) requests \\
$r$ & single tasking (observation) request \\
$L$ & set of all tiles \\
$l$ & single tile out of set of all tiles \\
$X$ & set of all collection opportunities \\
$X^S$ & set of all scheduled collection opportunities \\
$x$ & image collection opportunity \\
$t_0$ & start of scheduling horizon \\
$t_s$ & start of collect opportunity \\
$t_e$ & end of collect opportunity \\
$q_s$ & spacecraft attitude at start of collect opportunity \\
$q_e$ & spacecraft attitude at end of collect opportunity \\
$\mathcal{K}_r$ & request constraint function \\
$\mathcal{K}_s$ & scheduling constraint function \\
$\dot{\theta}_{\text{sc}}$ & spacecraft slew rate \\
$t_{\text{sc}}^{\text{settle}}$ & spacecraft maneuver settling time \\
$E_{\text{feas}}$ & edges in feasibility graph representation of scheduling problem \\
$E_{\text{infeas}}$ & edges in feasibility graph representation of scheduling problem \\
$\Delta$ & maximum degree of graph \\
$x^s$ & spacecraft $s$ associated with collect \\
$x^l$ & tile $l$ associated with collect
\end{longtable*}}

\section{Introduction}

The operation of Earth observing (EO) satellite systems requires efficient planning methods to coordinate activities of multiple assets. Currently operational constellations are comprised of tens~\cite{henely2019turning} to hundreds~\cite{boshuizen2014results,irisov2018recent} of spacecraft. These systems are relied upon to provide data for humanitarian, Earth science, climate science, and defense efforts~\cite{board2007earth}. They are expected to collect, update, and disseminate optical and radar imagery in a timely and responsive manner, enabling scientific studies and monitoring of changes due to human actors or natural phenomena. However, current and planned constellation sizes are beyond what can be practically managed using human operators alone. Automated task planning of on-orbit operations, in particular imaging activities, is needed to effectively manage Earth observing satellite systems. 

Easier access to space due to a proliferation of new launch vehicles combined with demonstrations of small satellites as cost-effective remote sensing platforms has led to the rise of a new generation of commercial remote sensing constellations~\cite{popkin2017commercial, stringham2019capella, sarda2018making}. These systems provide better temporal resolution from higher revisit rates, improved spatial resolution from operating in Low Earth Orbit, and greater robustness to single-point failures due to fractionated mission architectures~\cite{brown2006value}. The number of collection opportunities grows with the constellation size, allowing for greater total data collection, but also increasing the likelihood for duplicated work if activities are not properly coordinated. Duplication of effort results in unnecessary expenditure of limited on-board power and data resources, or missed collections of time-sensitive phenomena that would otherwise be possible given proper deconfliction. Furthermore, human-in-the-loop operations models for responsive (24-hour, 7-days a week) mission operations expect at least one dedicated mission planner working at all times for each satellite~\cite{wertz2011space}, making personnel costs alone prohibitive for constellations larger than a few spacecraft unless a high degree of automation is deployed.

The scheduling of imaging activities for spacecraft is known to be NP-complete~\cite{garey1979computers}. The planning space complexity grows exponentially with the number of collection opportunities. Multi-satellite scheduling increases the challenge by multiplying the number of collection opportunities with each added spacecraft. Constellation schedules must be replanned frequently in response to breaking events, commonly multiple times per-day, making long solution times operationally prohibitive. Consequently, various formulations and heuristics have been introduced over the years~\cite{hall1994maximizing,lemaitre2002selecting,bianchessi2005earth,beaumet2011feasibility}. Better planning approaches allow for more efficient and responsive operation of existing and future space assets, improving overall data collection throughput and enabling new science and business opportunities.

This paper introduces a scheduling technique based on finding maximum independent sets that can efficiently schedule and deconflict imaging activities for large constellations. The method relies upon posing the satellite planning problem as a graph where collection opportunities are represented by vertices and edges represent mutually exclusive collection opportunities. In this formulation the optimal schedule is any maximal independent set of vertices in the graph. We apply the ReduMIS algorithm, which combines local search, graph reduction, and evolutionary improvement to find the largest independent subset of nodes~\cite{lamm2017finding}.

Initial results show that the maximum independent set approach can simultaneously improve the number of scheduled collects and reduce solve time for large scheduling problems. Experiments show that for small problem sizes, those with 500 or fewer requested images, the maximum independent set approach is able to find optimal solutions with slightly worse runtime compared to baseline methods. For large problems, the method can improve solution quality by up to 8\% while simultaneously reducing scheduling time by 75\%. The method is tested on scenarios of up to 24 satellites and 10,000 requested observations. In addition, users can tune scheduling performance by setting the solver time limit, trading between solution quality and scheduling time.

The paper proceeds as follows. \Cref{sec:background} provides background on the satellite task planning problem as well as a brief review of previously studied approaches. \Cref{sec:modeling} introduces the modeling constructs and processing approach needed to specify the scheduling problem. \Cref{sec:mis} introduces graph formulations of the scheduling problem and the maximum independent set solution technique. \Cref{sec:results} provides experimental results, comparing the performance of the maximum independent set scheduling method to alternative scheduling methods. \Cref{sec:conclusion} concludes, summarizing the work and suggesting areas for future improvement.

\section{Background}
\label{sec:background}

The satellite task scheduling problem asks the planning agent to select  data collection opportunities that maximize an observation objective while simultaneously obeying system constraints. The inputs are a set of observation requests, the satellite trajectories, and a planning horizon. The most common objectives are maximizing  monetary value of collected data, timeliness of data return, or total number of collected images. The problem of assigning observations to single or multiple  spacecraft is equivalent to the precedence constrained scheduling problem, which has been shown to be NP-complete~\cite{garey1979computers}. As there are no known polynomial time solution algorithms, many approaches rely on heuristics, local search, or other approximate solution techniques. Several variations expand the problem scope by adding management of on-board resources (power and data), selection of communication times (scheduling ground contacts), or coordinating activities of multiple spacecraft. It is the latter extension that this paper addresses. 

The satellite task planning problem has been studied extensively for single, agile spacecraft, and more recently in the context of multi-satellite constellations.  One of the earliest computational solutions to problem was proposed by Hall and Magazine. They used heuristics and bounding techniques to find a feasible schedule~\cite{hall1994maximizing}. \Citeauthor{harrison1999task} study the problem of planning radar satellite imaging activities for a single pass over a region containing up to 50 desired observation locations. They use an enumeration method to construct possible schedules, scoring each on overall fitness. Since a full enumeration would be computationally intractable, the authors apply a tree pruning technique to the limit the size of the enumeration space~\cite{harrison1999task}. \Citeauthor{lemaitre2002selecting} study the problem for agile satellites in the PLEIADES program, in particular SPOT5. They provide dynamic programming, constraint programming, and heuristic local search algorithms for the problem~\cite{lemaitre2002selecting}. Due to the complexity of the problem, specifically the large size of the decision space encountered when considering long horizons, \citeauthor{lemaitre2002selecting} simplify the full problem by dividing it into separate smaller problems separated at half-orbit time boundaries. They find that this reduces the decision space to the point of computational tractability at the cost of being unable to coordinate activities across time boundaries, which leads to duplicated efforts or missed collection opportunities. The authors mitigate the inability to coordinate actions across half-orbit boundaries through the introduction of a discount factor to encourage collection of requests unique to each sub-problem. \Citeauthor{martin2002satellite} introduces a two horizon planning system used by the IKONOS commercial optical imaging satellite~\cite{martin2002satellite}. In this approach, planning occurs once daily over a long horizon (30-days) to incorporate future orders, and operators subsequently perform ad-hoc, short-horizon planning to incorporate last minute requests. The long planning horizon uses a network-flow approach while short-horizon planning is performed by human operators assisted by a proprietary dynamic programming planning tool. \Citeauthor{iacopino2012highly} applies ant colony optimization to solve the problem of planing of data collection for a single satellite~\cite{iacopino2012highly}.

In the context of multi-satellite constellations, \citeauthor{bianchessi2006mathematical} introduce a number of different approaches for the COSMO-Skymed constellation of three synthetic aperture radar satellites.  The first is a Lagrangian relaxation of an integer programming problem~\cite{bianchessi2006mathematical}. The schedule for each satellite is solved for separately with the solution order being chosen randomly. Tasking requests scheduled for one satellite are removed for the scheduling problem for subsequent satellites. \Citeauthor{bianchessi2007heuristic} also introduce a tabu search heuristic for the multi-satellite scheduling problem and use it to solve problems of up to 2,273 requested locations over a 26 orbit (approximately 39 hour) planning horizon~\cite{bianchessi2007heuristic}.  \Citeauthor{iacopino2013eo} extended their initial single-satellite ant colony evolutionary algorithm to successfully plan imaging and downlink opportunities for constellations of up to 16 satellites and 50 requested observations~\cite{iacopino2013eo, iacopino2013self}. \Citeauthor{augenstein2014optimal} studies the task planning problem for the SkyBox constellation of optical satellites. They introduce a graph-based planning method where data collection opportunities are represented by graph vertices and edges represent feasible transitions between collection opportunities~\cite{augenstein2014optimal}. The planning problem is solved by finding the longest weighted path through the resulting graph. The authors find a significant advantage of this technique is that it allows for predictable human intervention through the introduction of force-in or force-out schedule operations through the manipulation of edges in the graph. \Citeauthor{augenstein2016optimal} also consider a Mixed Integer Linear Programming (MILP) approach that accounts for data and power constraints. To solve the MILP quickly they first solve an approximation of the problem using dynamic programing and use that solution to warm-start the MILP, reducing overall computational effort and improving speed~\cite{augenstein2016optimal}. They find this approach performs better than the graph due to the improved coordination of activities which reduces the number of duplicated collections and conflicting actions. This approach is used to schedule the SkySat constellation in a scenario of up to 13 satellites and 7,000 requested locations over a 10 hour horizon. \Citeauthor{nag2018scheduling} use dynamic programming to maximizes the number of collection opportunities by finding the most efficient slew trajectory for imaging in satellite attitude domain. The authors apply their method to a four satellite constellation and up to 22,718 requested imaging locations~\cite{nag2018scheduling}. They handle inter-satellite coordination by applying the planning approach to each spacecraft separately but advancing time in parallel for all planners. A periodic resynchronization is performed to share collected images up to the current time between planning threads to reduce duplication of effort. The selection of the resynchronization frequency is highly dependent on the number of unique locations observed by each satellite. If the percentage of locations unique to any satellite is high, resynchronization can be performed less frequently due to the reduced likelihood of duplication of effort.

\section{Modeling Fundamentals}
\label{sec:modeling}

This article is concerned with the multi-agent version of the satellite scheduling problem\textemdash a single centralized agent must plan sensor collections of a set of observation requests $R$, for a constellation of satellites $S$, over a fixed planning horizon $H$. First, we need a way to represent the problem that can be later translated into actions the spacecraft can realize. There are multiple choices of how to model the problem that can map optimization outcomes into satellite actions. Some approaches define the action space to be line-of-sight trajectories in the satellite body frame~\cite{nag2018scheduling}. Others employ an action space comprised of discrete opportunities where imaging is abstracted as a single continuous action~\cite{augenstein2014optimal}. We adopt the latter approach and model the scheduling problem using discrete collection opportunities $X$, referred to as \textit{collects}. The scheduling problem then becomes finding a subset of collects $X^S \subseteq X$ such that the planning objective $\mathcal{O}$ obtains a maximum at $\mathcal{O}(X^S)$. This section describes the modeling approach and processing techniques used to setup and study the problem in subsequent sections.

\subsection{Tasking Request Constraints}

For each request $r \in R$, we define constraints on the imaging acquisition process. Many scientific applications require specific properties of the collected data and for image analysis techniques to provide accurate results. For example, detecting and extracting roadways from satellite imagery becomes significantly more challenging when the imagery is acquired from large off-nadir pointing angles~\cite{van2020road}. For a collect to provide valuable data, downstream analysis requirements must be accounted for during the collection planning process. We accomplish this by specifying collection constraints for each request, $K_r$. Request constraints are grouped into three categories: spatial, temporal, and imaging. Spatial constraints, specify what geographic extent to be captured. Temporal constraints specifying the period during which collections must occur. Imaging constraints determine the collection properties. \Cref{tab:request_constraints} provides a list of common constraints that apply to optical and radar imaging systems.

\begin{table}[htb]
\begin{center}
\caption{\label{tab:request_constraints}
Radar and optical imagery request constraints}
\makebox[\textwidth]{
\begin{tabular}{l l l}
\toprule
 Constraint & Description & Example \\	
 \midrule
 Request Validity Start & Start of validity period for image collection & 2021-07-01T00:00:00Z \\
 Request Validity End & Start of validity period for image collection & 2021-07-07T00:00:00Z \\
 Local Time & Required local time of collection & 0800 -- 1100 \\
 Look Angle & Angle between satellite line-of-sight and nadir vector & $25^{\circ} \leq \theta \leq 50^{\circ}$  \\
 Ground Range Resolution & Image resolution in the ground-range direction & $0.5 \text{m} \; \leq \text{GRR} \leq 1.5 \; \text{m}$ \\
 Azimuth Range Resolution & Image resolution in the azimuth direction & $0.5 \text{m} \; \leq \text{AZR} \leq 1.5 \; \text{m}$ \\
 NESZ & Radar noise equivalent sigma zero value & $NESZ < -15 \; \text{dB}$ \\
 Azimuth Angle & Azimuth angle of location with respect to satellite at time of acquisition & $225^{\circ} \leq az \leq 275^{\circ}$ \\
 Cloud Cover & Maximum percentage of scene obstructed by clouds & cloud\% $\leq 80\%$ \\
\bottomrule
\end{tabular}
}
\end{center}
\end{table}

\begin{figure}[htb]
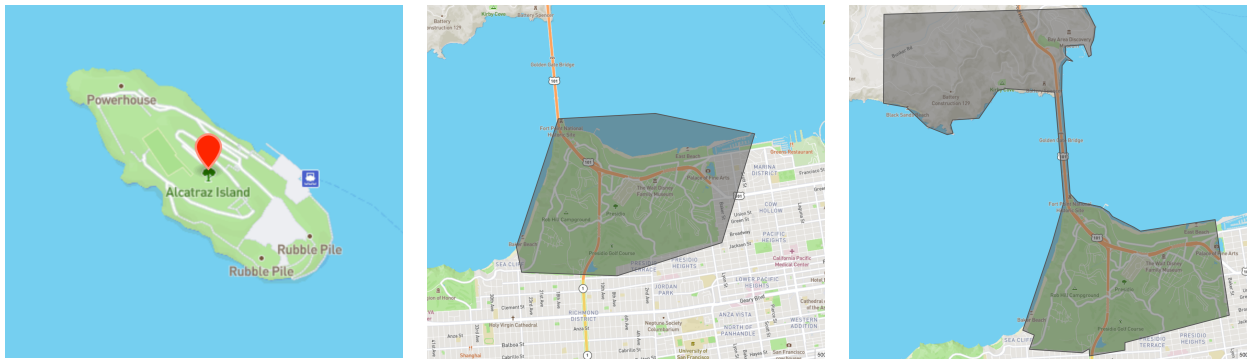

\include{request_geometries}
\caption{\label{fig:request_geometries}Possible request geometries. A simple point (left), convex polygon (middle), and non-convex polygon (right). All polygons are simple polygons, that do not interest themselves and contain no holes.}	
\end{figure}

A request's geometric boundary determines the minimal spatial extent, known as the area of interest, to be captured. This can be specified as a point or simple polygon geometry. \Cref{fig:request_geometries} provides examples of different possible request geometries. A point request is a single desired location and can be captured with a single image. A polygon represents an area where the entire area is desired for capture. If the area of a polygon is larger than the sensor field of view, multiple separate collects are required to fully capture the AOI. This leads to a subsequent challenge of decomposing the area into individually feasible sensor collections. Due to the continuous nature of the problem, there are infinitely many ways decomposition can occur. The decomposition is performed using a tessellation algorithm, which is discussed in \Cref{sec:request_processing}.

All non-spatial request constraints $\mathcal{K}_r$ are functions such that $k: T \times P \times L \times R \rightarrow \{0, 1\} \; \forall \; k \in \mathcal{K}_r$, where $T$ is a time in the planning horizon $[t_0, t_0 + H]$, $P$ is the satellite state vector in the Earth-fixed frame (position and velocity), $L$ is the set of tiles (discussed in \Cref{sec:request_processing}), and $R$ is the set of all requests. Request constraint functions map the inputs to a binary variable 1 to indicate that the collect falls in the valid domain, or it returns 0 if the constraint is not satisfied.

For many applications the image analysis results are only of value if the acquisition occurs within a specific time frame. Each request has validity window specified by a start and end time. In this work, we consider requests with a validity window coincident with the planning horizon, though in general, this need not be the case. A temporal constraint on the local time of acquisition also can be applied. Many applications, especially those related to human economic, cultural, and military activities align with the 24 hour day, therefore imaging at specific times allows for capture time-correlated phenomena.

Imaging constraints are used to specify the viewing geometry or sensor-dependent image properties of the collect. The primary constraint we consider is that of look angle $\theta$. The look angle is defined to be the angle between the satellite's nadir vector and the imaging payload's instantaneous line-of-site vector. Specific applications may pose requirements in terms of elevation angle or incidence angle that can be transformed into look angle constraints. It is important to note that the look angle constraint will have a significant impact on the number of collection opportunities found in any given horizon. Larger ranges of acceptable look angles will admit more possible collection opportunities, making deconfliction with other requests easier and increasing the likelihood of a request's inclusion in a schedule. This comes at the cost of greater scheduling complexity due to a larger action space. Other common imaging constraints include image resolution, image noise, and cloud coverage.


All constraints are applied with a logical AND so that every collect meets all request constraints. This approach has two benefits: first, all collects are guaranteed to meet the full set of desired collection properties, eliminating the possibility of collected imagery not meeting requirements after collection. It should be noted the strength of this guarantee rests on the ability to accurately model the constraints in request processing. The second benefit is that applying collection constraints during action space generation reduces the computational complexity of the scheduling problem. The upper bound on number of possible schedules given by brute-force enumeration is $2^{|X|}$ so eliminating unsuitable collections during request processing, prior to scheduling, can significantly reduce scheduling complexity and runtime.

\subsection{Action Space Determination}
\label{sec:request_processing}

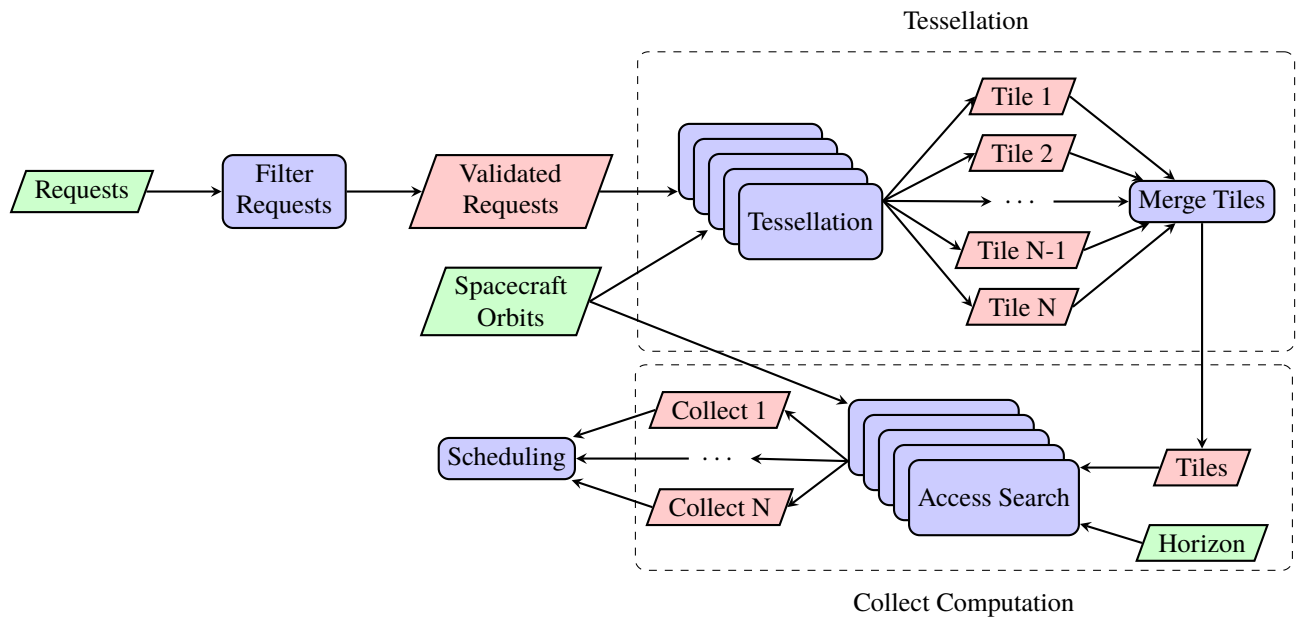
\begin{figure}[htb]
\begin{center}
\begin{tikzpicture}[square/.style={rectangle,minimum width=1cm,minimum
height=1cm,draw,thick,fill=blue!20}]

\node[trapezium, trapezium left angle=70, trapezium right angle=110,draw, thick, fill=green!20] (request) {Requests};

\node[rectangle, draw, thick, fill=blue!20, right=1cm of request,align=center, rounded corners, text width = 1.4 cm] (filter) {Filter Requests};

\node[trapezium, trapezium left angle=70, trapezium right angle=110,draw, thick, fill=red!20, right=1cm of filter, text width=1.75cm, align=center] (validated) {Validated Requests};

\node[square, above right=4mm and 2cm of validated.east, anchor=center, rounded corners] (tessellation1) {\phantom{Tessellation}};
\foreach \X [count=\Y] in {2,3,4} {\node[anchor=north west,below right=2mm and 2mm of tessellation\Y.north west,square, rounded corners] (tessellation\X){\phantom{Tessellation}};}
\node[anchor=north west,below right=2mm and 2mm of tessellation4.north west,square, rounded corners]  (tessellation5) {Tessellation};

\node[trapezium, trapezium left angle=70, trapezium right angle=110,draw, thick, fill=red!20, above right=0.5cm and 2cm of tessellation3, align=center] (tile1) {Tile 1};
\node[trapezium, trapezium left angle=70, trapezium right angle=110,draw, thick, fill=red!20, below=0.25cm of tile1, align=center] (tile2) {Tile 2};
\node[trapezium, trapezium left angle=70, trapezium right angle=110, below=0.25cm of tile2, align=center] (tile3) {\ldots};
\node[trapezium, trapezium left angle=70, trapezium right angle=110,draw, thick, fill=red!20, below=0.25cm of tile3, align=center] (tile4) {Tile N-1};
\node[trapezium, trapezium left angle=70, trapezium right angle=110,draw, thick, fill=red!20, below=0.25cm of tile4, align=center] (tile5) {Tile N};

\node[rectangle, draw, thick, fill=blue!20, right=1cm of tile3, align=center, rounded corners] (merge) {Merge Tiles};

\node[rectangle, draw, dashed, text width=8.5cm, text height=3.75cm, text centered, rounded corners, minimum height=2cm, left=0.75cm of tile3.center, anchor=center] (tessellation) {};
\node[text centered, above=0.5em of tessellation] (tessellation_text) {Tessellation};

\node[trapezium, trapezium left angle=70, trapezium right angle=110,draw, thick, fill=red!20, below=3cm of merge] (tiles) {Tiles};

\node[square, above left=4mm and 3cm of tiles.west, anchor=center, rounded corners] (access1) {\phantom{Access Search}};
\foreach \X [count=\Y] in {2,3,4} {\node[anchor=north west,below right=2mm and 2mm of access\Y.north west,square, rounded corners] (access\X){\phantom{Access Search}};}
\node[anchor=north west,below right=2mm and 2mm of access4.north west,square, rounded corners]  (access5) {Access Search};


\node[trapezium, trapezium left angle=70, trapezium right angle=110,draw, thick, fill=red!20, above left=0.0 and 1.85cm of access3, align=center] (collect1) {Collect 1};
\node[trapezium, trapezium left angle=70, trapezium right angle=110, below=0.25cm of collect1, align=center] (collect2) {\ldots};
\node[trapezium, trapezium left angle=70, trapezium right angle=110,draw, thick, fill=red!20, below=0.25cm of collect2, align=center] (collect3) {Collect N};

\node[rectangle, draw, dashed, text width=8.5cm, text height=2.5cm, text centered, rounded corners, minimum height=2cm, right=0cm of access3.center, anchor=center] (collect_computation) {};
\node[text centered, below=0.5em of collect_computation] (collect_computation_text) {Collect Computation};

\node[rectangle, draw, thick, fill=blue!20, left=1.5cm of collect2, align=center, rounded corners] (scheduling) {Scheduling};

\node[trapezium, trapezium left angle=70, trapezium right angle=110,draw, thick, fill=green!20, below=0.5cm of validated, text width=1.5cm, align=center] (spacecraft) {Spacecraft Orbits};

\node[trapezium, trapezium left angle=70, trapezium right angle=110,draw, thick, fill=green!20, below=0.5cm of tiles, align=center] (horizon) {Horizon};

\draw [->,thick,>=stealth] (request) -- (filter);
\draw [->,thick,>=stealth] (filter) -- (validated);
\draw [->,thick,>=stealth] (validated) -- ([xshift=-4mm]tessellation3.west);
\foreach \X [count=\Y] in {1,2,3,4,5} {\draw [->,thick,>=stealth] ([xshift=4mm,yshift=-1.25mm]tessellation3.east) -- (tile\X.west);}
\foreach \X [count=\Y] in {1,2,3,4,5} {\draw [->,thick,>=stealth] (tile\X.east) -- (merge);}
\draw [->,thick,>=stealth] (merge) -- (tiles);
\draw [->,thick,>=stealth] (tiles) -- ([xshift=4mm]access3.east);
\foreach \X [count=\Y] in {1,2,3} {\draw [->,thick,>=stealth] ([yshift=2mm]access1.south west) -- (collect\X.east);}
\foreach \X [count=\Y] in {1,2,3} {\draw [->,thick,>=stealth] (collect\X.west) -- (scheduling);}

\draw [->,thick,>=stealth] (spacecraft.east) -- (tessellation3.south west);
\draw [->,thick,>=stealth] (spacecraft.east) -- (access1);
\draw [->,thick,>=stealth] (horizon.west) -- (access5);

\end{tikzpicture}
\end{center}
\caption{Request processing pipeline. Inputs are shown in green, computational processing in blue, and data outputs in red. The tessellation and access search steps can occur in parallel with computation divided between worker nodes.}
\label{fig:request_processing_pipeline}
\end{figure}

Prior to deciding which collects to include in the schedule, we must first find all possible collection opportunities $X$, for all $R$, over the horizon $H$. When $|R|$ is large, the problem of finding the collect opportunities becomes computationally intensive, though one well suited to parallelization. The request processing pipeline, shown in \Cref{fig:request_processing_pipeline}, takes inputs of $(R,S,H)$ and outputs the set of all collects $X$. There are four major steps:
\begin{enumerate}
	\item Filtering requests with contradictory constraints or ill-defined geometries.
	\item Tessellating each request based on each satellites' orbit parameters to determine a set of tiles. 
	\item Merging any overlapping tiles.
	\item Performing an access search to find all collection opportunities that satisfy request constraints.
\end{enumerate}

The first step in processing requests is to validate that the request geometry passes basic consistency checks to ensure they can be processed. All requests must be points or simple polygons. The polygons can be convex or non-convex, but they cannot have any self-intersections or interior segments. Filtering is used to remove requests with contradictory request constraints, as any such request will not yield any collects and therefore needs not be processed. The filtering also applies heuristic-checks to speed subsequent processing by removing requests known to be infeasible by definition, e.g., the request location lies above the constellation's highest visited latitude.

The next step of processing is tessellation, where we decomposing the requested spatial extent into individual tiles that can be captured by the sensor payload in a single action. For point geometries this process is straight forward as imaging the scene can be done in a single collect. For area requests the process is more complicated as there are an infinite number of possible tilings that could cover the requested area. Therefore, the process is somewhat arbitrary and application dependent, however the approach should guarantee complete coverage of the AOI and minimize the amount of excess collected area. Tessellation can be done using a flat-plane geometry where the algorithm assumes all points exist on a flat-plane. This leads to a simpler algorithm, but can lead to gaps in coverage for large AOIs when the effect of Earth's curvature must be accounted for. We use an approach based on spherical geometry that can tessellate areas of arbitrary size. 
The output of this step is the set of all tiles $L$. The tiles generated by request $r$ are denoted $L_r$.

%

The final step in processing is finding all collect opportunities that satisfy all of the request constraints, $\mathcal{K}_r$, for each tile of the request. The output of this step is the set of collect opportunities $X$. Collects are a tuple of information
\begin{equation}
	x \equiv (t_s, t_e, q_s, q_e, s, l, r)
\end{equation}

Each collect has an associated start time $t_s$ and end time $t_e$. The spacecraft must continuously maintain a specific attitude for the duration of the collection. For staring imaging modes this means continuously slewing the spacecraft to track a point on Earth's surface. For strip imaging modes the attitude fixed in the body frame and natural orbital motion is used to scan the ground in the along-track direction. For both type of imaging modes, there is a required initial attitude $q_s$ that the spacecraft must be in at $t_s$ to initiate the collection as well as a final attitude $q_e$ at end time $t_e$. Each collect is also associated with a specific spacecraft $s$, tile $l$, and request $r$. This set of information is required for applying scheduling constraints to a problem formulation.

Finding all collects is done using a binary search algorithm presented in \Cref{alg:binary_search}. The algorithm first performs a initial coarse search to find times where all collect constraints are satisfied and then preforms a recursive binary search around each coarse time to determine the precise collection time boundaries. The inputs to the algorithms are a request $r$, it's associated tiles $L_r$, and spacecraft properties and orbit information $s$. We use NORAD Two Line Element (TLE) sets and associated SGP4 propagator for orbit prediction in this work~\cite{vallado2006revisiting}, however the orbit prediction model could be exchanged for high-fidelity dynamics model if more accuracy is needed. The collect search is executed in parallel for all combinations spacecraft, requests, and tiles. The collects found from this process become the action space of the scheduling problem, where an action is the decision to take a collect or not.

\begin{algorithm}[htb]
\caption{Find Collect Tiles}
\label{alg:binary_search}
\begin{algorithmic}[1]
\Function{TileCollectSearch}{$t_0,t_f,s,l,r,dt,tol$}
\State $X_l \gets []$
\State $t \gets t_0$
\While{$t < t_f$}
	\State $p \gets \textsc{PredictECEFState}(t,s)$
	\If{$\textsc{CheckAccessConstraints}(t,p,l,r) = 1$}
		\State $t_s \gets \textsc{FindCollectBoundary}(t,s,l,r,-dt,tol)$
		\State $t_e \gets \textsc{FindCollectBoundary}(t,s,l,r,dt,tol)$
		\State $q_s \gets \textsc{ComputeSCAttitude}(t_s,p,l)$
		\State $q_e \gets \textsc{ComputeSCAttitude}(t_e,p,l)$
		\State $X_l \gets \textsc{Append}(X_l, (t_s,t_e,q_s,q_e,s,l,r))$
		\State $t \gets t + t_{macro}$
	\Else
		\State $t \gets t + t_{micro}$
	\EndIf
\EndWhile
\State \Return $X_l$
\EndFunction
\Function{CheckAccessConstraints}{$t,p,l,r$}
	\State $v \gets \textsc{True}$
	\For{$k \in K_r$}
		\State $v \gets v \land k(t,p,l,r)$
		\If{$v = \textsc{False}$}
			\State \Return $\textsc{False}$
		\EndIf
	\EndFor
	\State \Return $\textsc{True}$
\EndFunction
\Function{FindCollectBoundary}{$t,s,l,r,dt$}
	\If{$dt < tol$}
		\State \Return $t$
	\Else
		\State $p \gets \textsc{PredictECEFState}(t,s)$
		\State $v \gets \textsc{CheckAccessConstraints}(t,p,l,r)$
		\While $\textsc{CheckAccessConstraints}(t,p,l,r) = v$
			\State $t \gets t + dt$
			\State $p \gets \textsc{PredictECEFState}(t,s)$
		\EndWhile
		\State $dt \gets -\textsc{sign}(dt) \times \frac{dt}{2}$
		\State \Return $\textsc{FindCollectBoundary}(t,s,l,r,dt,tol)$
	\EndIf
\EndFunction
\end{algorithmic}
\end{algorithm}


\subsection{Scheduling Constraints}

After finding the set of possible collection opportunities $X$, we need a way to  determine which collects can be simultaneously included in the schedule and which pairs of collects are mutually exclusive. We define a set of scheduling constraints functions $\mathcal{K}_s$ such that $k: X \times X \rightarrow \{0,1\} \; \forall \; k \in \mathcal{K}_s$. Each constraint function takes as input two distinct collects and maps it to an indicator variable with a value of 1 if both collects can be accommodated and 0 if not. We consider two scheduling constraints in this work\textemdash a spacecraft agility constraint $k_{agility}$ and a collect repetition constraint $k_{repetition}$. The spacecraft agility constraint enforces that there is enough time to reorient the spacecraft between collects. The repetition constraint ensures that duplicated collections are not admitted into the schedule.

The most important constraint is the spacecraft agility constraint. Without the agility constraint schedules would not be physically realizable. As discussed in the request processing section, the spacecraft must reach a specific initial attitude at the start of every collect $q_s$ and end the collect in attitude $q_e$. Between every collect pair, the spacecraft must slew between $q_e$ of the preceding collect and reach $q_s$ of the subsequent collect prior to the the start of imaging. The spacecraft agility constraint $k_{agility}(x_i,x_j)$ determines whether a slew is feasible between collects $x_i$ and $x_j$ by checking if there is enough time to complete the maneuver. The constraint function is defined to be
\begin{equation}
k_{agility}(x_i, x_j) = \begin{cases} 
      1 & \text{if } f_a(q_e^i, q_s^j) \leq t_s^j - t_e^i \\
      0 & \text{otherwise}
   \end{cases}
\end{equation}
where the spacecraft agility model $f_a(q_1,q_2)$ returns the time required for the spacecraft to move between attitudes $q_1$ and $q_2$. We apply a simple angular velocity slew model that assumes the spacecraft slews at a constant rate $\dot{\theta}_{\text{sc}}$ and settles in fixed duration $t_{\text{sc}}^{\text{settle}}$. The agility model is
\begin{equation}
	f_a(q_1,q_2) = \frac{q_1 \cdot q_2^{-1}}{\dot{\theta}_{\text{sc}}} + t_{\text{sc}}^{\text{settle}} = t_{slew}
\end{equation}
While this model does not capture the full satellite attitude dynamics model it is fast to evaluate numerically, and proper selection of $\dot{\theta}_{\text{sc}}$ and $t_{\text{sc}}^{\text{settle}}$ allows the model to bound worst-case slew times.

The second constraint is the repetition constraint. The repetition constraint eliminates the possibility of duplicated collection effort by checking if two collects are of the same tile. The constraint is defined as
\begin{equation}
k_{repetition}(x_i, x_j) = \begin{cases} 
	1 & \text{if } x^l_i \neq x^l_j \\
    0 & \text{otherwise}
\end{cases}
\end{equation}

While not applied in this work, the force-in and force-out constraints of Augenstein~\cite{augenstein2014optimal} are of practical value for supporting manual intervention in scheduling. However, they do not directly fit into the above constraint definition framework. It is possible retain their effect by filtering $X$ based on the desired force-in/force-out operation. A force-out constraint requires a specific collect \textit{not} be taken. This can be accomplish by simply removing all force-out collects from $X$ prior to scheduling
\begin{equation}
	X \gets X \setminus X_{FORCE-OUT}
\end{equation}
The force-in constraint requires that any collect forced-in \textit{must} be scheduled. This can be accomplished by removing all collects that conflict with any forced-in collect as the could not be included in the schedule anyways. The force-in filtered set of scheduling collects is
\begin{equation}
	X \gets \{ x \; | \; k(x,x_f) = 0 \; \forall \; k \in \mathcal{K}_s , \; x_f \in X_{FORCE-IN} \}
\end{equation}
Enforcing the force-in constraint prior to, as opposed to during, scheduling helps reduce the scheduling problem complexity.

\subsection{Objective Function}


Finally, we define the scheduling objective function $\mathcal{O}$ which is used to quantify the value of different schedules. This is accomplished by assigning each request a weight $w_i^r$ that represents the reward for capturing imagery of a request. The total value of a schedule simply becomes the total of value of all collected imagery
\begin{equation}
	\mathcal{O}(X) = \sum_{x_i \in X} w^r_i
\end{equation}
If all requests are equally valued, $w^r = 1 \; \forall \; r \in R$, the objective becomes
\begin{equation}
	\mathcal{O}(X) = |X|
\label{eqn:scheduling_objective}
\end{equation}
which is simply maximizing the total number of collects. 

\section{Graph Model and Maximum Independent Set Formulation}
\label{sec:mis}

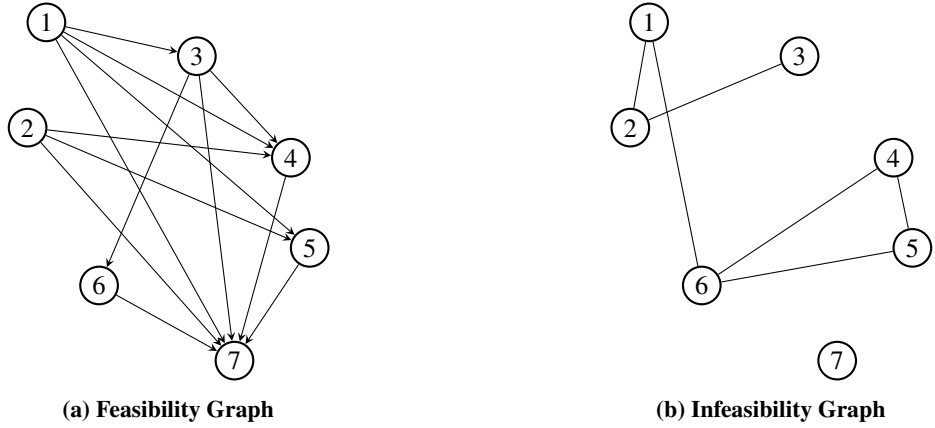
\begin{figure}[htb]
\begin{center}

\begin{subfigure}{.48\textwidth}
\centering
\begin{tikzpicture}[every node/.style={draw=black,thick,circle,inner sep=0pt}]
\node[circle,draw,minimum size=0.5cm] at (0,0) (n1) {$1$};
\node[circle,draw,minimum size=0.5cm] at (2.,-0.45) (n3) {$3$};
\node[circle,draw,minimum size=0.5cm] at (3.25,-1.8) (n4) {$4$};
\node[circle,draw,minimum size=0.5cm] at (3.5,-3) (n5) {$5$};
\node[circle,draw,minimum size=0.5cm] at (-0.25,-1.4) (n2) {$2$};
\node[circle,draw,minimum size=0.5cm] at (0.7,-3.5) (n6) {$6$};
\node[circle,draw,minimum size=0.5cm] at (2.5,-4.5) (n7) {$7$};
\draw [->,>=stealth] (n1) -- (n3);
\draw [->,>=stealth] (n1) -- (n4);
\draw [->,>=stealth] (n1) -- (n5);
\draw [->,>=stealth] (n1) -- (n7);
\draw [->,>=stealth] (n2) -- (n4);
\draw [->,>=stealth] (n2) -- (n5);
\draw [->,>=stealth] (n2) -- (n7);
\draw [->,>=stealth] (n3) -- (n7);
\draw [->,>=stealth] (n3) -- (n6);
\draw [->,>=stealth] (n3) -- (n4); 
\draw [->,>=stealth] (n4) -- (n7);
\draw [->,>=stealth] (n5) -- (n7); 
\draw [->,>=stealth] (n6) -- (n7);
\end{tikzpicture}
\caption{Feasibility Graph}
\end{subfigure}
\begin{subfigure}{.48\textwidth}
\centering
\begin{tikzpicture}[every node/.style={draw=black,thick,circle,inner sep=0pt}]
\node[circle,draw,minimum size=0.5cm] at (0,0) (n1) {$1$};
\node[circle,draw,minimum size=0.5cm] at (2.,-0.45) (n3) {$3$};
\node[circle,draw,minimum size=0.5cm] at (3.25,-1.8) (n4) {$4$};
\node[circle,draw,minimum size=0.5cm] at (3.5,-3) (n5) {$5$};
\node[circle,draw,minimum size=0.5cm] at (-0.25,-1.4) (n2) {$2$};
\node[circle,draw,minimum size=0.5cm] at (0.7,-3.5) (n6) {$6$};
\node[circle,draw,minimum size=0.5cm] at (2.5,-4.5) (n7) {$7$};
\draw (n1) -- (n2);
\draw (n1) -- (n6);
\draw (n2) -- (n3);
\draw (n4) -- (n5);
\draw (n4) -- (n6);
\draw (n5) -- (n6);
\end{tikzpicture}
\caption{Infeasibility Graph}
\end{subfigure}%

\end{center}
\caption{Both interpretations of same scheduling problem. The feasibility interpretation (left) represents the problem as an directed acyclic graph. The infeasibility graph (right) is an undirected graph. The edges in the two graphs are complimentary.}
\label{fig:graph_formulation}	
\end{figure}

We can formulate the scheduling problem as solving the maximum independent set problem for a graph. The collects and scheduling constraints can be represented as a graph $G = (V,E)$. The graph vertices correspond to collects, $V \equiv X$, and edges correspond to constraints. There are two ways to construct the graph, as shown in \Cref{fig:graph_formulation}. The first method applies a \textit{feasibility} interpretation to the problem, where edges represent feasible transitions between collects; an edge is present when both agility and repetition constraints are satisfied. The second is the \textit{infeasibility} interpretation where edges in the graph represent pairs of collects that are mutually exclusive. The maximum independent set formulation is based upon the infeasibility interpretation, though we will introduce and discuss both to provide additional understanding of the satellite scheduling problem and context for the advantages of the maximum independent set approach.

\subsection{Graph Interpretations of the Scheduling Problem}

The feasibility interpretation of the problem takes the view that after each collection the planning agent selects the next collect to take out of the entire set of possible collects. The satellite performs the selected collection, advancing to the corresponding vertex in the graph, and the process repeats. Because the agent can only move forward in time, result is a directed acyclic graph (DAG). The scheduling problem then becomes deciding which edge (action) maximizes the total planning reward over the entire horizon. In this approach, the edges in the graph correspond to pairs of collects where all scheduling constraints are satisfied. That is
\begin{equation}
	E_{\text{feas}} = \{(x_i,x_j) \; | \; k_{agility}(x_i,x_j) \land k_{repetition}(x_i,x_j) = 1 \}
\end{equation}
The feasibility view is used by many planning methods including the heuristics rules of \citeauthor{bianchessi2007heuristic}~\cite{bianchessi2007heuristic}, the longest weighted path approach of \citeauthor{augenstein2014optimal}~\cite{augenstein2014optimal}, and the Markov Decision Process formulation of \citeauthor{eddyMarkov2020}~\cite{eddyMarkov2020}.

The infeasibility interpretation takes the opposite approach. The edges correspond to any pair of collects where any scheduling constraints are not met. That is
\begin{equation}
	E_{\text{infeas}} = \{(x_i,x_j) \; | \; k_{agility}(x_i,x_j) \lor k_{repetition}(x_i,x_j) = 0 \; \}
\end{equation}
This approach results in an undirected graph, as the planning agent is not moving along edges between vertices, but is instead selecting the most valuable subset of collects that are not mutually exclusive. The structure underlying the infeasibility view forms the basis for mixed-integer linear programming approaches~\cite{augenstein2016optimal,nag2018scheduling} in addition to forming the basis for the maximum independent set solution we use in this paper.

\subsection{Graph Interpretation Properties}
\label{sec:graph_interpretation}

There are a few observations about the satellite scheduling problem we can make at this point. The first is that the feasibility and infeasibility interpretations are complimentary. Consider a feasibility graph $G_{\text{feas}}$. For any two vertices in the graph $(x_i,x_j)$ there will be an edge present if all scheduling constraints are satisfied. If a single constraint is not satisfied, the collects are mutually exclusive, and no edge will exist in $E_{\text{feas}}$. However, because there is at least one unsatisfied constraint, the edge between $(x_i,x_j)$ \textit{must} exist in $E_{\text{infeas}}$. Conversely, if an edge is present in the infeasibility graph, there cannot be an edge in the feasibility graph. If we define the set of all possible edges for a graph to be $E$, the following must hold true for the single satellite scheduling problem
\begin{align}
	E_{\text{feas}} \cap E_{\text{infeas}} & = \emptyset \\
	E_{\text{infeas}} & = E \setminus E_{\text{feas}} \label{eqn:infeas_transform} \\
	E_{\text{feas}} & = E \setminus E_{\text{infeas}} \label{eqn:feas_transform} \\
	|E_{\text{feas}} \cup E_{\text{infeas}}| & = \frac{1}{2}|X|\left(|X| - 1\right) = |E|
\end{align}
\Cref{eqn:infeas_transform,eqn:feas_transform} make it possible to transform between the feasibility and infeasibility interpretations of the problem simply by writing the problem in graph form and taking the complementary set of edges. The directionality of edges can be determined by comparing the start time of two collects, with any edges in the feasibility DAG originating from the earlier and ending at the later vertex.

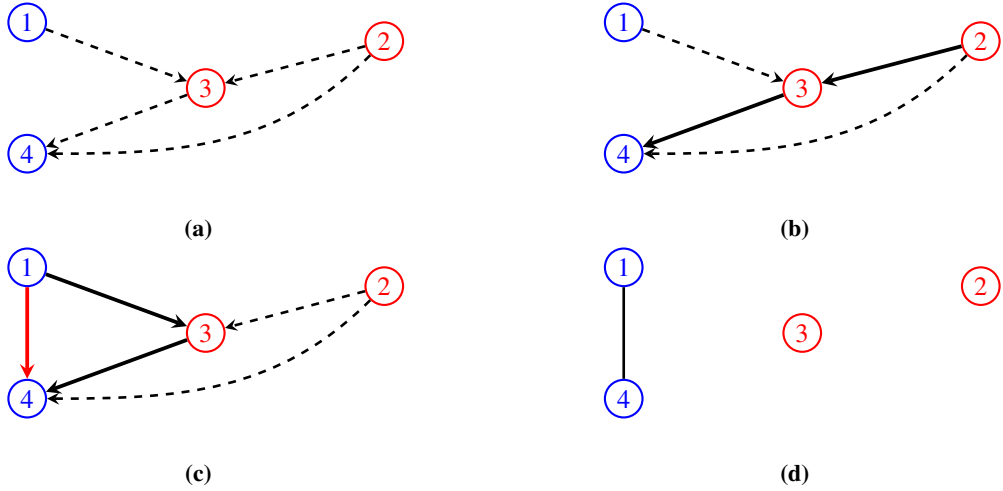
\begin{figure}[htb]
\begin{center}
	
\begin{subfigure}{.475\textwidth}
\begin{center}
\begin{tikzpicture}[every node/.style={draw=black,thick,circle,inner sep=0pt}]
\node[circle,draw,color=blue,minimum size=0.5cm] at (0,0) (n1) {$1$};
\node[circle,draw,color=red,minimum size=0.5cm, below right= 0.5cm and 2cm of n1] (n3) {$3$};
\node[circle,draw,color=red,minimum size=0.5cm, above right= 0.25cm and 2cm of n3] (n2) {$2$};
\node[circle,draw,color=blue,minimum size=0.5cm, below left= 0.5cm and 2cm of n3] (n4) {$4$};

\draw [->,>=stealth,line width=0.35mm,dashed] (n1) -- (n3);
\draw [->,>=stealth,line width=0.35mm,dashed] (n2) -- (n3); 
\draw [->,>=stealth,line width=0.35mm,dashed] (n3) -- (n4);
\draw [->,>=stealth,line width=0.35mm,dashed] (n2) to[out=225,in=0] (n4);

\end{tikzpicture}
\end{center}
\caption{\phantom{-}}
\end{subfigure}
\begin{subfigure}{.475\textwidth}
\begin{center}
\begin{tikzpicture}[every node/.style={draw=black,thick,circle,inner sep=0pt}]
\node[circle,draw,color=blue,minimum size=0.5cm] at (0,0) (n1) {$1$};
\node[circle,draw,color=red,minimum size=0.5cm, below right= 0.5cm and 2cm of n1] (n3) {$3$};
\node[circle,draw,color=red,minimum size=0.5cm, above right= 0.25cm and 2cm of n3] (n2) {$2$};
\node[circle,draw,color=blue,minimum size=0.5cm, below left= 0.5cm and 2cm of n3] (n4) {$4$};

\draw [->,>=stealth,line width=0.35mm,dashed] (n1) -- (n3);
\draw [->,>=stealth,line width=0.5mm] (n2) -- (n3); 
\draw [->,>=stealth,line width=0.5mm] (n3) -- (n4);
\draw [->,>=stealth,line width=0.35mm,dashed] (n2) to[out=225,in=0] (n4);

\end{tikzpicture}
\end{center}
\caption{\phantom{-}}
\end{subfigure}%

\begin{subfigure}{.475\textwidth}
\begin{center}
\begin{tikzpicture}[every node/.style={draw=black,thick,circle,inner sep=0pt}]
\node[circle,draw,color=blue,minimum size=0.5cm] at (0,0) (n1) {$1$};
\node[circle,draw,color=red,minimum size=0.5cm, below right= 0.5cm and 2cm of n1] (n3) {$3$};
\node[circle,draw,color=red,minimum size=0.5cm, above right= 0.25cm and 2cm of n3] (n2) {$2$};
\node[circle,draw,color=blue,minimum size=0.5cm, below left= 0.5cm and 2cm of n3] (n4) {$4$};

\draw [->,>=stealth,line width=0.5mm] (n1) -- (n3);
\draw [->,>=stealth,line width=0.35mm,dashed] (n2) -- (n3); 
\draw [->,>=stealth,line width=0.5mm] (n3) -- (n4);
\draw [->,>=stealth,line width=0.5mm,color=red] (n1) -- (n4);
\draw [->,>=stealth,line width=0.35mm,dashed] (n2) to[out=225,in=0] (n4);

\end{tikzpicture}
\end{center}
\caption{\phantom{-}}
\end{subfigure}
\begin{subfigure}{.475\textwidth}
\begin{center}
\begin{tikzpicture}[every node/.style={draw=black,thick,circle,inner sep=0pt}]
\node[circle,draw,color=blue,minimum size=0.5cm] at (0,0) (n1) {$1$};
\node[circle,draw,color=red,minimum size=0.5cm, below right= 0.5cm and 2cm of n1] (n3) {$3$};
\node[circle,draw,color=red,minimum size=0.5cm, above right= 0.25cm and 2cm of n3] (n2) {$2$};
\node[circle,draw,color=blue,minimum size=0.5cm, below left= 0.5cm and 2cm of n3] (n4) {$4$};

\draw [line width=0.35mm] (n1) -- (n4);

\end{tikzpicture}
\end{center}
\caption{\phantom{-}}
\end{subfigure}%

\end{center}
\caption{Naive feasibility interpretation of 2 satellite, 2 collect schedule problem where all inter-satellite transitions are allowed. Collects numbers 1--4 and colored by satellite as blue or red (a). One valid schedule with valid transitions across satellites (b). Invalid schedule as the path results in a direct transition for the blue satellite not contained in its single-satellite sub-graph (c). Infeasibility interpretation where no contradiction exists (d).}
\label{fig:feasibility_contradictions}	
\end{figure}

In the multi-satellite scheduling problem the relationship between the feasibility and infeasibility interpretations becomes more complicated. When more than one satellite is present, the feasibility interpretation cannot model transitions between satellites without admitting contradictions. To see why consider a simple scheduling scenario of 2 satellites with 2 collects each, shown in \Cref{fig:feasibility_contradictions}. Intuitively, the agility and attitude of one spacecraft cannot affect the ability of different spacecraft to take collect collect, therefore all transitions between collects where $x_i^s \neq x_j^s$ should be part of the graph. However, this can admit situations where agility constraints of an individual spacecraft, can be circumvented by transition between spacecraft around an absent edge. If this occurs, the resulting schedule would not be feasible from the single-satellite perspective as it would require an infeasible transition to realize.

The feasibility interpretation can be extended to multi-satellite scheduling if transitions between satellites are disallowed. The set of feasible edges for the feasibility graph becomes
\begin{equation}
	E_{\text{feas}} = \{(x_i,x_j) \; | \; k_{agility}(x_i,x_j) \land k_{repetition}(x_i,x_j) \land (x_i^s = x_j^s) = 1 \}
\end{equation}
Transformations between the multi-satellite feasibility and infeasibility graphs remains possibly so long as the edge relationships are updated to account for the lack inter-satellite transitions $E_{\text{is}}$ in the feasibility interpretation.
\begin{align}
	E_{\text{is}} & = \{(x_i,x_j) \; | \; x_i^s \neq x_j^s \} \\
	E_{\text{feas}} \cap E_{\text{infeas}} \cap E_{\text{is}} & = \emptyset \\
	E_{\text{infeas}} & = E \setminus (E_{\text{feas}} \cap E_{\text{is}}) \\
	E_{\text{feas}} & = E \setminus (E_{\text{infeas}} \cap E_{\text{is}})  \\
	|E_{\text{feas}} \cup E_{\text{infeas}} \cap E_{\text{is}}| & = \frac{1}{2}|X|\left(|X| - 1\right) = |E|
\end{align}

One consequence of the multi-satellite feasibility interpretation being unable to support inter-satellite edges is that solvers that work by traversing the feasibility graph do not automatically share collection history (state information) required to prevent repeat collects occurring with different spacecraft. Multi-satellite scheduling algorithms must solve for each satellite separately in serial or solve for each satellite in parallel, which makes global coordination of actions across all satellites over the entire time domain impossible. Instead feasibility graph approaches have to employ discount factors or periodic resynchronization steps to indirectly coordinate activities, as done in~\cite{nag2018scheduling}. The infeasibility view is not affected by these challenges. 

\begin{figure}[htb]
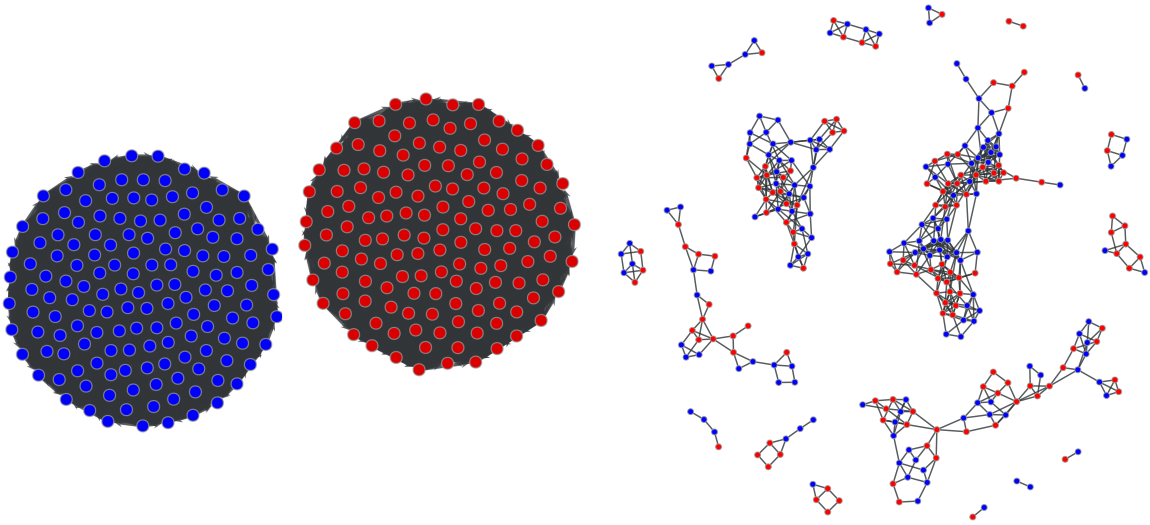

\include{real_graph}
\caption{Resulting graph structure of feasibility (left) and infeasibility interpretation (right) of the satellite scheduling problem for a 2 satellite, 100 request, 287 collect scenario. Feasibility graph appearance of vertices on black background is due to the underlying graph being almost fully dense.}
\label{fig:real_graph_structure}
\end{figure}

Another observation is that, for most planning horizons, we expect that the feasibility view to result in a dense graph due to the nature of the agility constraint. Consider a satellite with $\dot{\theta}_{\text{sc}} = $ \SI[per-mode=symbol]{1}{\degree\per\second} and $t_{\text{sc}}^{\text{settle}} = $ \SI{15}{\second}, values representative of current small satellite attitude control systems. A \SI{180}{\degree} slew maneuver, the worst-case reorientation, will take at most \SI{195}{\second}. This means that any collect pair $(x_i,x_j)$ with $t^i_e + 195 < t^j_s$ will always be feasible with regards to the agility constraint. As most planning horizons are a few hours long or more, the vast majority of future collects are feasible transitions. It is only the short time around a collection where other collects might not be reachable due to the agility constraint. This property can be seen in \Cref{fig:real_graph_structure}. Therefore, for most satellite scheduling problems we expect the feasibility graph to be dense because most transitions between nodes are possible.

\begin{table}[htb]
\caption{Comparison of edge density for single satellite scheduling problems with 100 to 10,000 point requests with agility and repetition scheduling constraints applied.}
\begin{center}

\begin{subfigure}{.475\textwidth}
\begin{center}
\caption{Feasibility View}

\begin{tabular}{r r r r r}
\toprule
Requests & Vertices & Edges & Density \\
\midrule
100 & 138 & 9,221 & 97.55 \% \\
200 & 284 & 39,217 & 97.59 \% \\
500 & 707 & 243,507 & 97.57 \% \\
1000 & 1,434 & 1,006,215 & 97.93 \% \\
2000 & 2,855 & 4,004,666 & 98.30 \% \\
5000 & 7,154 & 25,218,482 & 98.56 \% \\
10000 & 14,221 & 97,309,964 & 96.24 \% \\
\bottomrule
\end{tabular}
\end{center}

\end{subfigure}
\begin{subfigure}{.475\textwidth}
\begin{center}
\caption{Infeasibility View}

\begin{tabular}{r r r r r}
\toprule
Requests & Vertices & Edges & Density \\
\midrule
100 & 138 & 232 & 2.45 \% \\
200 & 284 & 969 & 2.41 \% \\
500 & 707 & 6,064 & 2.43 \% \\
1000 & 1,434 & 21,246 & 2.07 \% \\
2000 & 2,855 & 69,419 & 1.70 \% \\
5000 & 7,154 & 367,799 & 1.44 \% \\
10000 & 14,221 & 3,801,346 & 3.76 \% \\
\bottomrule
\end{tabular}
\end{center}

\end{subfigure}%
\end{center}
\label{tab:edge_density}	
\end{table}

To confirm this expectation, we computed the collects and edges for test scenarios of up to 10,000 distinct point requests over a 24 hour horizon. The results, shown in \Cref{tab:edge_density}, confirm that the feasibility view produces a dense graph with a 96.24\% or more of possible edges present across all problem sizes considered. Similarly, we find a correspondingly sparse infeasibility graph as expected from the complimentary nature of the two graph interpretations. Since all scheduling algorithms need to iterate over edges to either consider feasible transitions or eliminate infeasible choices, we expect infeasibility-based approaches to be more efficient because there are fewer edges.

\subsection{Maximum Independent Set Algorithms}

The maximum independent set (MIS) solution to the satellite scheduling problem is based on the intuition that if the problem is represented as an infeasibility graph, the optimal schedule is simply the largest set of vertices for which there are no common edges. This problem is equivalent to the maximum independent set problem in graph theory. The maximum independent set problem requires finding the largest set of vertices $I \subseteq$ V , such that no vertices in $I$ are adjacent to one another. The resulting set is called a maximum independent set. The solution to the maximum independent set problem is not necessarily unique as there may exist multiple independent sets of equal cardinality. Any such set is called a maximal independent set. Recent algorithms can efficiently search for independent sets in large sparse graphs. Techniques for solving the MIS problem include local search, evolution, and recursive reduction.

The first technique for finding maximal independent sets is that of swaps. A swap is a local search algorithm that maintains a candidate independent set of vertices $\mathcal{S}$ and iteratively improves it by adding, removing, and exchanging vertices. \Citeauthor{andrade2012fast} generalize the notion as $(j,k)$-swaps, where the $j$ vertices are removed from the solution and $k$ are added~\cite{andrade2012fast}. When $j < k$ the swap becomes an improvement, referred to as a $k$-improvement, as it increases size of the candidate independent set. \Citeauthor{andrade2012fast} introduce fast algorithms that either fina  a valid 1, 2, or 3-improvement or proving no $k$-degree improvement is possible.  The algorithm for 1-improvements is constant time, 2-improvements linear time $O(|E|)$, and 3-improvements algorithm is at worst quadratic $O(|E|\Delta)$ where $\Delta$ is the maximum degree of the graph. Since the 2-improvement algorithm of \citeauthor{andrade2012fast} has linear time complexity in the number of edges for any vertex it is extremely well suited for the satellite scheduling problem as the problem is expected to be extremely sparse.

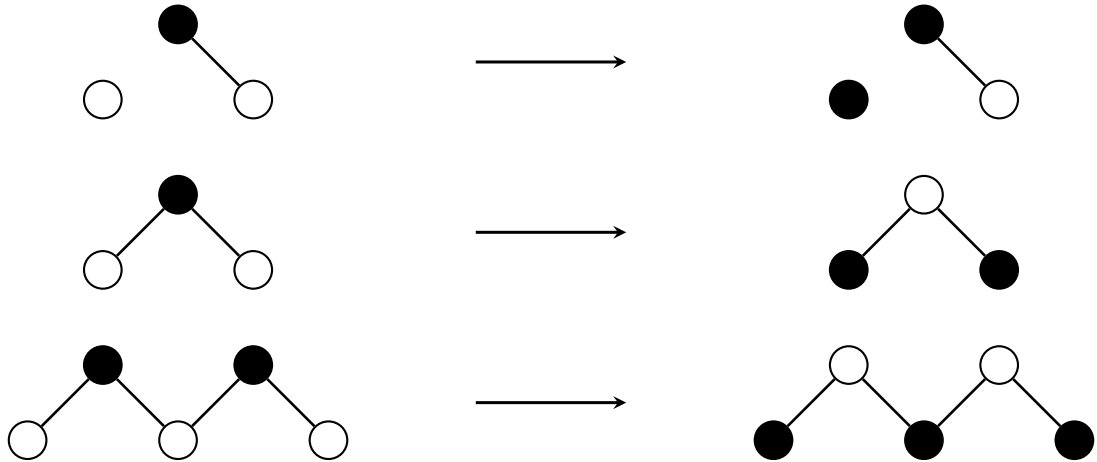
\begin{figure}[htb]
\begin{center}

\begin{subfigure}{\textwidth}
\centering

\begin{minipage}{0.39\textwidth}
\begin{center}
\begin{tikzpicture}[every node/.style={draw=black,thick,circle,inner sep=0pt}]

\node[circle,draw,minimum size=0.5cm,fill=black] at (-1,0.5) (n11) {};
\node[circle,draw,minimum size=0.5cm] at (-2,-0.5) (n13) {};
\node[circle,draw,minimum size=0.5cm] at (0,-0.5) (n14) {};
\draw [line width=0.35mm] (n11) -- (n14);

\end{tikzpicture}
\end{center}
\end{minipage}
\begin{minipage}{0.2\textwidth}
\begin{center}
\begin{tikzpicture}[every node/.style={draw=black,thick,circle,inner sep=0pt}]

\draw [->,>=stealth,line width=0.4mm] (-1,0) -- (1,0);

\end{tikzpicture}
\end{center}
\end{minipage}
\begin{minipage}{0.39\textwidth}
\begin{center}
\begin{tikzpicture}[every node/.style={draw=black,thick,circle,inner sep=0pt}]

\node[circle,draw,minimum size=0.5cm,fill=black] at (-1,0.5) (n11) {};
\node[circle,draw,minimum size=0.5cm,fill=black] at (-2,-0.5) (n13) {};
\node[circle,draw,minimum size=0.5cm] at (0,-0.5) (n14) {};
\draw [line width=0.35mm] (n11) -- (n14);

\end{tikzpicture}
\end{center}
\end{minipage}
\end{subfigure}

\vspace{2em}

\begin{subfigure}{\textwidth}
\centering

\begin{minipage}{0.39\textwidth}
\begin{center}
\begin{tikzpicture}[every node/.style={draw=black,thick,circle,inner sep=0pt}]

\node[circle,draw,minimum size=0.5cm,fill=black] at (-1,0.5) (n11) {};
\node[circle,draw,minimum size=0.5cm] at (-2,-0.5) (n13) {};
\node[circle,draw,minimum size=0.5cm] at (0,-0.5) (n14) {};
\draw [line width=0.35mm] (n11) -- (n13);
\draw [line width=0.35mm] (n11) -- (n14);

\end{tikzpicture}
\end{center}
\end{minipage}
\begin{minipage}{0.2\textwidth}
\begin{center}
\begin{tikzpicture}[every node/.style={draw=black,thick,circle,inner sep=0pt}]

\draw [->,>=stealth,line width=0.4mm] (-1,0) -- (1,0);

\end{tikzpicture}
\end{center}
\end{minipage}
\begin{minipage}{0.39\textwidth}
\begin{center}
\begin{tikzpicture}[every node/.style={draw=black,thick,circle,inner sep=0pt}]

\node[circle,draw,minimum size=0.5cm] at (-1,0.5) (n11) {};
\node[circle,draw,minimum size=0.5cm,fill=black] at (-2,-0.5) (n13) {};
\node[circle,draw,minimum size=0.5cm,fill=black] at (0,-0.5) (n14) {};
\draw [line width=0.35mm] (n11) -- (n13);
\draw [line width=0.35mm] (n11) -- (n14);

\end{tikzpicture}
\end{center}
\end{minipage}
\end{subfigure}%

\vspace{2em}

\begin{subfigure}{\textwidth}
\centering

\begin{minipage}{0.39\textwidth}
\begin{center}
\begin{tikzpicture}[every node/.style={draw=black,thick,circle,inner sep=0pt}]

\node[circle,draw,minimum size=0.5cm,fill=black] at (-1,0.5) (n11) {};
\node[circle,draw,minimum size=0.5cm,fill=black] at (1,0.5) (n12) {};
\node[circle,draw,minimum size=0.5cm] at (-2,-0.5) (n13) {};
\node[circle,draw,minimum size=0.5cm] at (0,-0.5) (n14) {};
\node[circle,draw,minimum size=0.5cm] at (2,-0.5) (n15) {};
\draw [line width=0.35mm] (n11) -- (n13);
\draw [line width=0.35mm] (n11) -- (n14);
\draw [line width=0.35mm] (n12) -- (n14);
\draw [line width=0.35mm] (n12) -- (n15);

\end{tikzpicture}
\end{center}
\end{minipage}
\begin{minipage}{0.2\textwidth}
\begin{center}
\begin{tikzpicture}[every node/.style={draw=black,thick,circle,inner sep=0pt}]

\draw [->,>=stealth,line width=0.4mm] (-1,0) -- (1,0);
\end{tikzpicture}
\end{center}
\end{minipage}
\begin{minipage}{0.39\textwidth}
\begin{center}
\begin{tikzpicture}[every node/.style={draw=black,thick,circle,inner sep=0pt}]

\node[circle,draw,minimum size=0.5cm] at (-1,0.5) (n11) {};
\node[circle,draw,minimum size=0.5cm] at (1,0.5) (n12) {};
\node[circle,draw,minimum size=0.5cm,fill=black] at (-2,-0.5) (n13) {};
\node[circle,draw,minimum size=0.5cm,fill=black] at (0,-0.5) (n14) {};
\node[circle,draw,minimum size=0.5cm,fill=black] at (2,-0.5) (n15) {};
\draw [line width=0.35mm] (n11) -- (n13);
\draw [line width=0.35mm] (n11) -- (n14);
\draw [line width=0.35mm] (n12) -- (n14);
\draw [line width=0.35mm] (n12) -- (n15);

\end{tikzpicture}
\end{center}
\end{minipage}
\end{subfigure}%

\end{center}
\caption{Graphical representation of 1, 2, and 3-improvements. Filled-in vertices indicate inclusion in the candidate independent set. 1-improvements (top) directly insert a vertex. 2-improvements (middle) directly remove one vertex and add two. 3-improvements (bottom) remove 2 vertices and add 3.}
\label{fig:k_improvements}	
\end{figure}

\begin{figure}[htb]
\begin{center}
\includegraphics[width=0.75\textwidth]{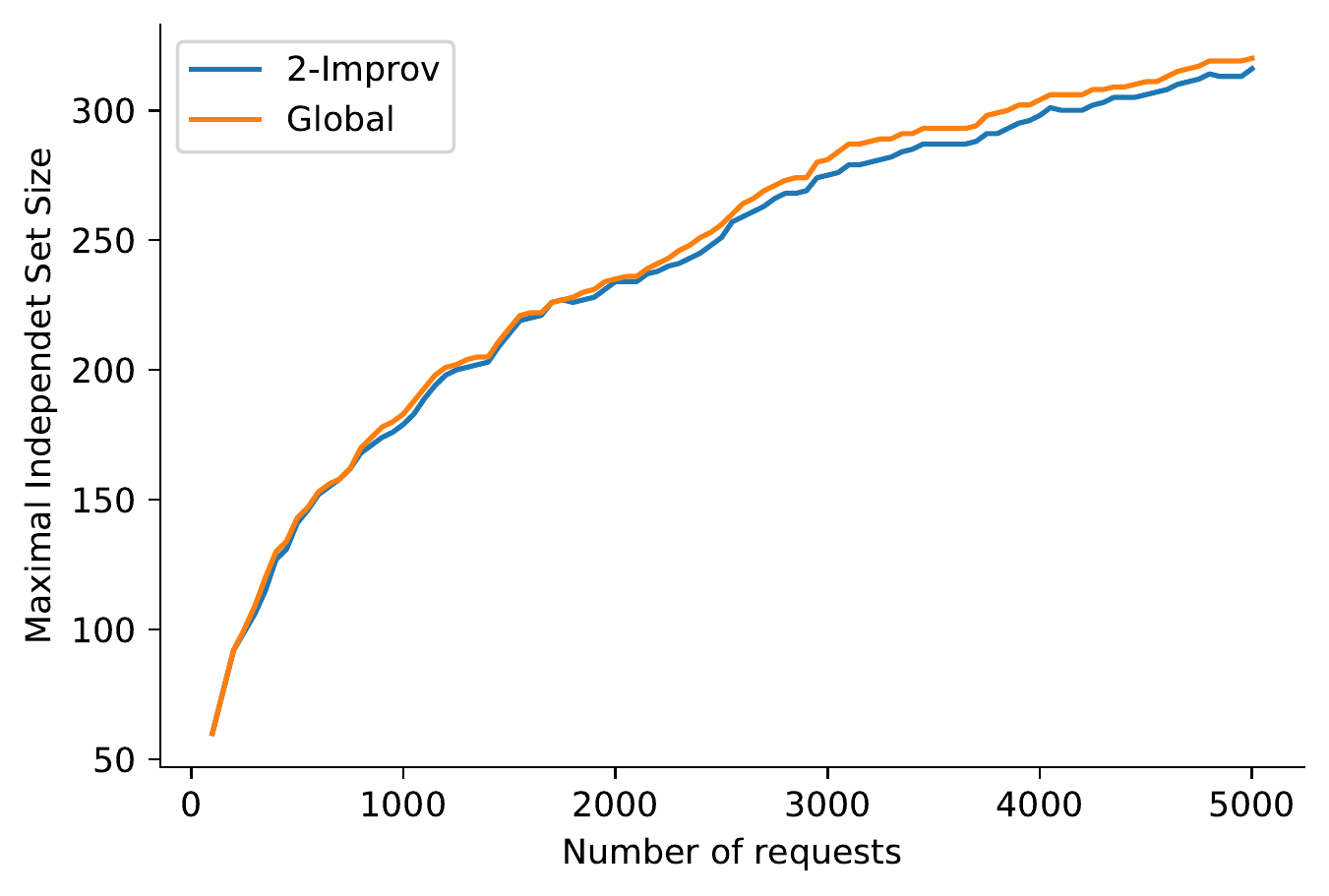}
\end{center}
\caption{Comparison of one-shot 2-improvement local search outcomes and exact global optimum found through Mixed Integer Linear Programming for simulated satellite scheduling problems of 100 to 5000 requests.}
\label{fig:2_improvement_suboptimality}
\end{figure}

Although the 2-improvement technique is extremely efficient it does not always find the optimal solution as it is a local search technique. \Cref{fig:2_improvement_suboptimality} shows a comparison between the satellite scheduling problem solved with a 2-improvement algorithm and the exact solution to the problem found using a MILP formulation and a duality-gap requirement of 0. This comparison shows that the 2-improvement solution returns near-optimal, but not optimal solutions therefore we also seek to apply additional solution techniques to improve solution optimality.

\begin{figure}[htb]
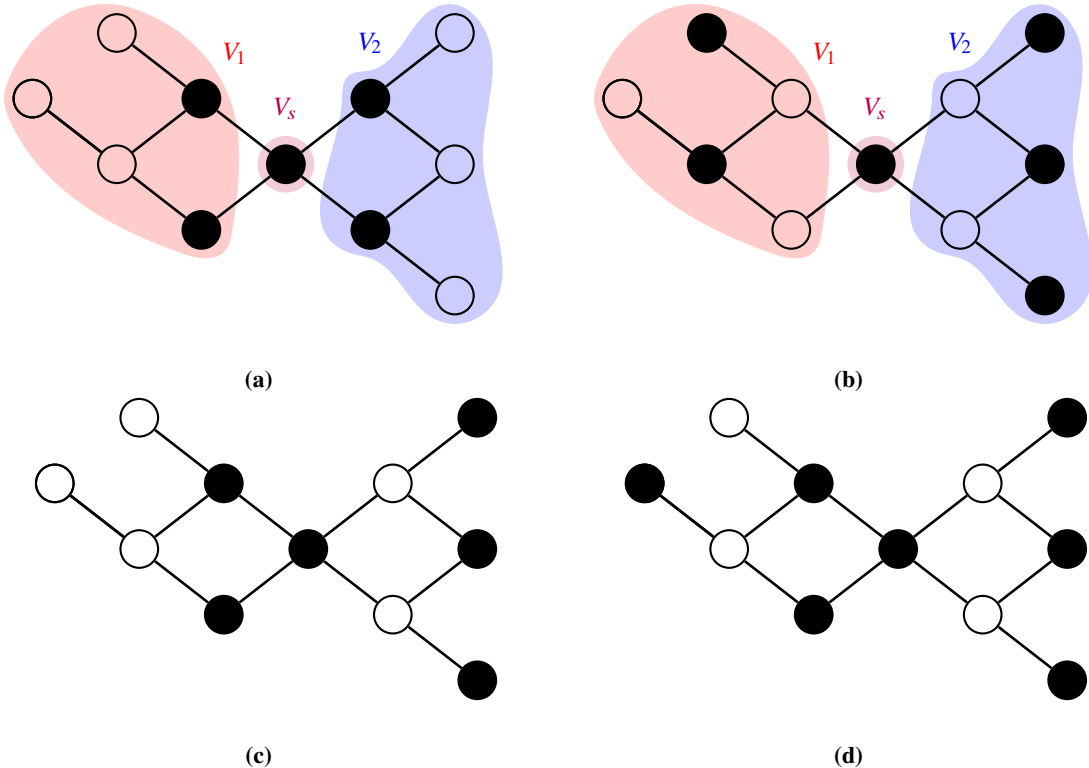

\include{evo_mis}
\caption{Graphical evolutionary maximum independent set algorithm. Two parent independent sets (a) and (b) are combined based on selected partitions $V_1$ and $V_2$ (c). Local search algorithm is then applied to create optimized child independent set (d).}
\label{fig:evo_mis}	
\end{figure}

One evolutionary algorithm introduced by \citeauthor{lamm2015graph}~\cite{lamm2015graph}, reproduced in \Cref{alg:evomis} and illustrated in \Cref{fig:evo_mis}, allows for finding larger independent sets than the $k$-improvement method alone. The evolutionary algorithm works by partitioning the graph using a 2-way node separator $V = V_1 \cup V_2 \cup V_s$, then producing a child node $O$ by combining members of the exiting independent set population $I_1$ and $I_2$ with the separator sets. A child can be produced using set combinations of $O = (V_1 \cap I_1) \cup (V_2 \cap I_2)$ or $O = (V_1 \cap I_2) \cup (V2 \cap I_1)$. In the final step of the algorithm child nodes are individually improved using the 2-improvement local search. The stopping criteria for the algorithm may be either total runtime or number of evolutionary generations that fail to produce improved offspring.

\begin{algorithm}[htb]
\caption{Evolutionary Maximum Independent Set Search}
\label{alg:evomis}
\begin{algorithmic}[1]
\Function{EvolveIndependentSet}{$V,E,t_{max}$}
	\State Create initial population $P$
	\While{Stop criteria not met}
		\State Select $I_1$, $I_2$ from $P$
		\State Combine $I_1$ and $I_2$ to create offspring $O$
		\State Improve $O$ using 2-improvement search
		\State Evict $O$'s closest dominated individual in $P$
	\EndWhile
	\State \Return Maximum size set in $P$
\EndFunction
\end{algorithmic}
\end{algorithm}

This evolutionary algorithm can be extended by combining it with a suite of graph reduction methods to improve the efficiency of the solver for sparse graphs in an algorithm called ReduMIS. The algorithm seeks to speed up the overall solution time using a branch-solve-technique, where sub-graphs of the of the problem are recursively generated and solved quickly using the evolutionary and the local search algorithms above. For a full discussion of the kernelization techniques used to reduce the graph refer to~\cite{lamm2017finding}. While still an inexact algorithm, not guaranteed to find the maximum independent set, the method has been tested on large webpage-relationship graphs of up to 1,382,908 vertices and was able to reliably find optimal solutions independently confirmed using slower, exact methods. It is the ReduMIS algorithm, which combines the maximum independent set solution techniques discussed above, that we apply to solve the satellite scheduling problem. 


\section{Results and Analysis}
\label{sec:results}

We compare the maximum independent set solution methods against two approaches previously used to solve the satellite task scheduling problem: a longest-weighted-path graph traversal approach and a MILP formulation.

\subsection{Tasking Request Definitions}

\begin{figure}[htb]
\begin{center}
\includegraphics[width=\textwidth]{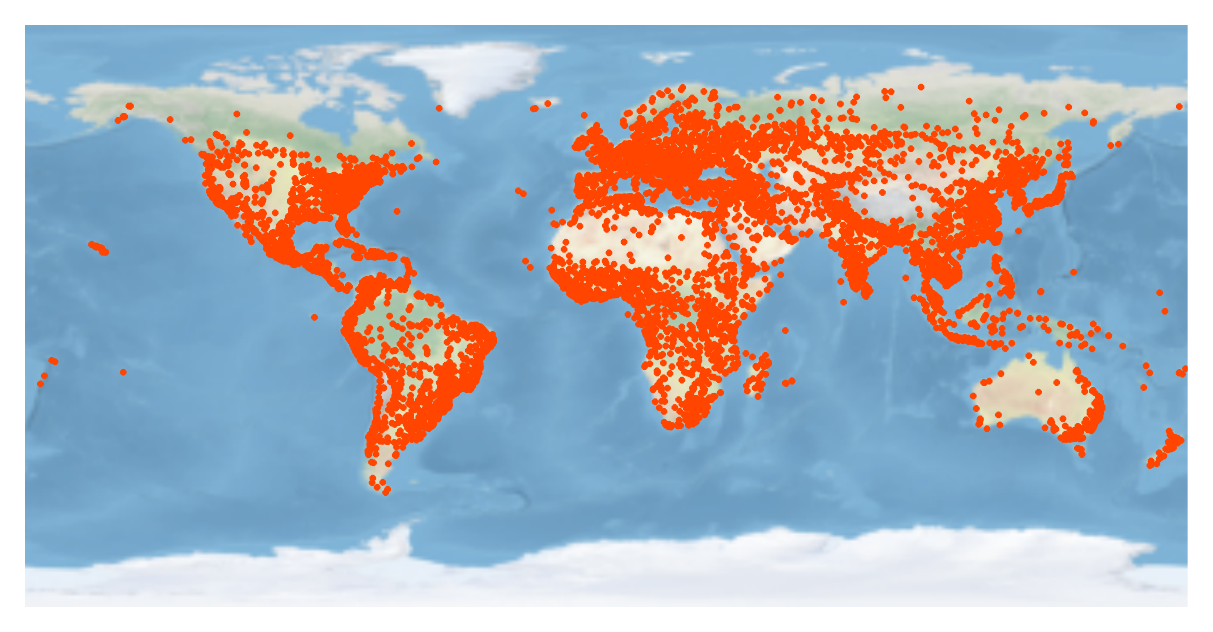}
\end{center}
\caption{10,000 request locations derived from Simplemaps' World Cities Database. Population centers serve as an approximation for imagery demand.}
\label{fig:requests}
\end{figure}

To evaluate the performance of the planning methods we need a representative set of tasking requests for an Earth observation constellation. Tasking requests received by commercial images companies are both proprietary and sensitive information so no request datasets based on commercial demand are currently publicly available, therefore a suitable substitute must be found. In the past, others have used LandSat WRS-2 grid points as request locations, however the WRS-2 grid covers the entire planet at an evenly spaced interval and extends between \SI{80.02}{\degree} latitude north and south. This covers a significant portion of the poles as well as large uninhabited regions, such as the Siberian tundra, Sahara desert, and Antarctica. An additional concern is that regular grid spacings are not representative of regions with high population density. Since agility constraints, and therefore underlying graph structure, are largely determined by the geographic distribution of requests we believe that variably spaced requests derived from population centers provide scenario more closely aligned to request sets encountered in practice. In our experiments, we use the locations of human population centers as point targets requests with high correlation to human-induced activity. We used the largest 10,000 population centers from Simplemap's World Cities Database\footnote{Data set distributed under the Creative Commons Attribution License 4.0 and accessible at https://simplemaps.com/data/world-cities.} to build the set of tasking requests. The locations of these 10,000 requests is shown in \Cref{fig:requests}.

\subsection{Experimental Results}

We consider scheduling constellations of 4, 6, 12, and 24 total satellites deployed in a Walker delta distribution over a 24 hour planning horizon. Represented in Walker notation the planes and phasing of the considered constellations are: 4/4/1, 6/2/1, 12/4/1, and 24/8/1. All spacecraft orbits are simulated using in \SI{500}{\kilo\meter}, circular polar orbits. We also consider planning activities for the 18 satellite SkySat constellation, the largest currently operational Earth observation constellation. The Skysat orbital geometry is taken from the Space-Track TLE database.\footnote{Accessed on July 23, 2020.} For the spacecraft agility model we assume that all spacecraft have a \SI[per-mode=symbol]{1}{\degree\per\second} constant slew rate and \SI{15}{\second} settling time.  All simulations were run on a workstation with a dual 14-core 2.6 GHz Intel E5-2690 processor and 128 GB of memory. For each constellation size, we consider sub-problems of scheduling 100, 200, 500, 1,000, 2,000, 5,000, 7,000, and 10,000 requests. We solve the satellite scheduling problem for each combination of  constellation and request set using the MIS ReduMIS algorithm, as well as the graph traversal and MILP formulations from \citeauthor{eddyMarkov2020}~\cite{eddyMarkov2020}. In additional to these baseline solutions, we also tested two alternate configurations of the MIS algorithm\textemdash one that is sent a termination signal after 60 seconds of solve time and a second that is sent the signal after 10 seconds. Finally, we also test a MILP formulation where the solver immediately returns after finding the first feasible solution. All solvers are configured to send a termination signal after \SI{900}{\second} of  execution time. How the termination signal is handled varies between solvers. Some solvers take significantly longer to return the active solution once requested to terminate. All schedules are validated by applying secondary checks against the agility and repetition constraints.

\begin{table}
\caption{Results for Walker constellations}
\begin{center}
\makebox[\textwidth]{\small
\begin{tabular}{c r r r r r r r r r r r r r r}
\toprule
\multicolumn{3}{c}{Scenario} & \multicolumn{2}{c}{MIS}  & \multicolumn{2}{c}{MIS (\SI{60}{\second})} & \multicolumn{2}{c}{MIS (\SI{10}{\second})} & \multicolumn{2}{c}{MILP} & \multicolumn{2}{c}{MILP ($1\text{st}$ Soln.)}&  \multicolumn{2}{c}{Graph} \\
\cmidrule{1-3} \cmidrule{4-5} \cmidrule{6-7} \cmidrule{8-9} \cmidrule{10-11} \cmidrule{12-13} \cmidrule{14-15}
$|S|$ & \multicolumn{1}{c}{$|R|$} & \multicolumn{1}{c}{$|X|$} & $|X^S|$ & $t_{solve}$ & $|X^S|$ & $t_{solve}$ & $|X^S|$ & $t_{solve}$ & $|X^S|$ & $t_{solve}$ & $|X^S|$ & $t_{solve}$ & $|X^S|$ & $t_{solve}$\\
\midrule
4 & 100 & 565 & \textbf{100} & 11.66 & \textbf{100} & 12.1 & \textbf{100} & 10.08 & \textbf{100} & 0.22 & 99 & 0.21 & \textbf{100} & \textbf{0.03} \\
4 & 200 & 1139 & \textbf{187} & 46.93 & \textbf{187} & 47.51 & \textbf{187} & 10.16 & \textbf{187} & 0.35 & 181 & 0.31 & 182 & \textbf{0.13} \\
4 & 500 & 2913 & 384 & 220.96 & 384 & 60.54 & 384 & 10.23 & \textbf{385} & 1.46 & 356 & \textbf{0.95} & 359 & 1.13 \\
4 & 1000 & 5856 & \textbf{600} & 553.43 & \textbf{600} & 60.39 & \textbf{600} & 10.39 & \textbf{600} & 4.47 & 541 & \textbf{2.97} & 532 & 5.85 \\
4 & 2000 & 11760 & \textbf{845} & 900.75 & 844 & 60.91 & 844 & 10.64 & \textbf{845} & 21.76 & 766 & \textbf{9.5} & 777 & 28.18 \\
4 & 5000 & 29518 & \textbf{1199} & 901.94 & 1194 & 65.12 & 1186 & \textbf{13.18} & \textbf{1199} & 842.48 & 1105 & 47.91 & 1129 & 206.24 \\
4 & 7000 & 41406 & \textbf{1316} & 915.44 & 1307 & 79.26 & 1303 & \textbf{33.73} & 1261 & 1035.85 & 1231 & 133.94 & 1249 & 430.67 \\
4 & 10000 & 59356 & \textbf{1413} & 987.21 & 1402 & 174.27 & 1376 & \textbf{91.41} & 1343 & 1450.78 & 1337 & 522.96 & 1051 & 901.04 \\
\midrule
6 & 100 & 855 & \textbf{100} & 3.1 & \textbf{100} & 3.03 & \textbf{100} & 3.02 & \textbf{100} & 0.37 & \textbf{100} & 0.37 & \textbf{100} & \textbf{0.03} \\
6 & 200 & 1721 & \textbf{200} & 37.38 & \textbf{200} & 37.77 & \textbf{200} & 10.27 & \textbf{200} & 0.53 & 197 & 0.51 & 194 & \textbf{0.14} \\
6 & 500 & 4376 & \textbf{443} & 274.42 & \textbf{443} & 60.37 & \textbf{443} & 10.46 & \textbf{443} & 2.15 & 419 & 1.54 & 422 & \textbf{1.26} \\
6 & 1000 & 8828 & \textbf{755} & 900.32 & 753 & 60.72 & 751 & 10.7 & \textbf{755} & 10.98 & 678 & \textbf{4.65} & 687 & 7.22 \\
6 & 2000 & 17718 & 1162 & 901.09 & 1159 & 60.87 & 1154 & \textbf{11.12} & \textbf{1163} & 118.55 & 1045 & 13.92 & 1061 & 37.33 \\
6 & 5000 & 44631 & \textbf{1706} & 903.59 & 1693 & 62.85 & 1679 & \textbf{16.92} & 1612 & 970.99 & 1561 & 77.69 & 1584 & 287.78 \\
6 & 7000 & 63068 & \textbf{1870} & 915.25 & 1855 & 75.17 & 1855 & \textbf{54.58} & 1779 & 1125.1 & 1745 & 220.33 & 1782 & 615.51 \\
6 & 10000 & 90777 & \textbf{2016} & 1006.6 & 2003 & 288.05 & 1952 & \textbf{138.61} & 1900 & 2031.2 & 1896 & 784.39 & 1061 & 901.23 \\
\midrule
12 & 100 & 1700 & \textbf{100} & 0.46 & \textbf{100} & 0.39 & \textbf{100} & 0.35 & \textbf{100} & 0.46 & \textbf{100} & 0.43 & \textbf{100} & \textbf{0.03} \\
12 & 200 & 3431 & \textbf{200} & 18.33 & \textbf{200} & 18.13 & \textbf{200} & 10.42 & \textbf{200} & 0.76 & \textbf{200} & 0.72 & \textbf{200} & \textbf{0.13} \\
12 & 500 & 8699 & \textbf{500} & 493.06 & \textbf{500} & 60.64 & \textbf{500} & 10.51 & \textbf{500} & 3.91 & 487 & 2.77 & 489 & \textbf{1.35} \\
12 & 1000 & 17574 & \textbf{948} & 900.67 & 946 & 60.97 & 945 & 10.86 & \textbf{948} & 25.12 & 894 & \textbf{8.31} & 901 & 8.32 \\
12 & 2000 & 35279 & \textbf{1730} & 901.3 & 1720 & 61.26 & 1712 & \textbf{12.03} & 1727 & 927.67 & 1569 & 27.33 & 1581 & 46.98 \\
12 & 5000 & 88993 & \textbf{2933} & 905.15 & 2890 & 64.24 & 2884 & \textbf{40.05} & 2654 & 1051.69 & 2654 & 153.59 & 2689 & 435.71 \\
12 & 7000 & 125317 & \textbf{3289} & 938.18 & 3243 & 90.92 & 3232 & \textbf{87.02} & 3018 & 1326.45 & 3017 & 427.15 & 2689 & 920.97 \\
12 & 10000 & 180008 & \textbf{3570} & 1018.22 & 3508 & 340.23 & 3418 & \textbf{240.3} & 3308 & 2660.79 & 3302 & 1746.45 & 1061 & 938.77 \\
\midrule
24 & 100 & 3405 & \textbf{100} & 0.86 & \textbf{100} & 0.77 & \textbf{100} & 0.77 & \textbf{100} & 0.65 & \textbf{100} & 0.59 & \textbf{100} & \textbf{0.04} \\
24 & 200 & 6884 & \textbf{200} & 1.22 & \textbf{200} & 1.29 & \textbf{200} & 1.27 & \textbf{200} & 1.35 & \textbf{200} & 1.28 & \textbf{200} & \textbf{0.15} \\
24 & 500 & 17430 & \textbf{500} & 900.86 & \textbf{500} & 62.1 & \textbf{500} & 16.75 & \textbf{500} & 6.61 & \textbf{500} & 5.29 & \textbf{500} & \textbf{1.4} \\
24 & 1000 & 35168 & \textbf{1000} & 903.85 & \textbf{1000} & 62.85 & \textbf{1000} & 22.16 & \textbf{1000} & 22.26 & \textbf{1000} & 17.46 & \textbf{1000} & \textbf{8.57} \\
24 & 2000 & 70581 & \textbf{2000} & 903.56 & 1999 & 63.55 & 1999 & \textbf{35.65} & \textbf{2000} & 129.67 & 1943 & 55.46 & 1951 & 50.76 \\
24 & 5000 & 177895 & \textbf{4319} & 909.6 & 4268 & \textbf{96.26} & 4268 & 97.46 & 3962 & 1208.93 & 3960 & 307.62 & 4025 & 552.77 \\
24 & 7000 & 250326 & \textbf{5019} & 936.47 & 4990 & 229.73 & 4946 & \textbf{216.81} & 4624 & 1759.61 & 4624 & 798.09 & 2481 & 900.31 \\
24 & 10000 & 359170 & \textbf{5566} & 1153.33 & 5482 & 654.22 & 5323 & \textbf{552.65} & 5153 & 4576.03 & 5153 & 3549.46 & 1061 & 933.45 \\
\bottomrule
\end{tabular}
}
\end{center}
\label{tab:walker_scheduling_results}
\end{table}

\begin{table}
\caption{Results for Skysat constellations}
\begin{center}
\makebox[\textwidth]{\small
\begin{tabular}{c r r r r r r r r r r r r r r}
\toprule
\multicolumn{3}{c}{Scenario} & \multicolumn{2}{c}{MIS}  & \multicolumn{2}{c}{MIS (\SI{60}{\second})} & \multicolumn{2}{c}{MIS (\SI{10}{\second})} & \multicolumn{2}{c}{MILP} & \multicolumn{2}{c}{MILP ($1\text{st}$ Soln.)} & \multicolumn{2}{c}{Graph} \\
\cmidrule{1-3} \cmidrule{4-5} \cmidrule{6-7} \cmidrule{8-9} \cmidrule{10-11} \cmidrule{12-13} \cmidrule{14-15}
$|S|$ & \multicolumn{1}{c}{$|R|$} & \multicolumn{1}{c}{$|X|$} & $|X^S|$ & $t_{solve}$ & $|X^S|$ & $t_{solve}$ & $|X^S|$ & $t_{solve}$ & $|X^S|$ & $t_{solve}$ & $|X^S|$ & $t_{solve}$ & $|X^S|$ & $t_{solve}$\\
\midrule
18 & 100 & 2388 & \textbf{100} & 8.25 & \textbf{100} & 8.2 & \textbf{100} & 8.26 & \textbf{100} & 0.4 & \textbf{100} & 0.39 & \textbf{100} & \textbf{0.04} \\
18 & 200 & 4860 & \textbf{200} & 31.22 & \textbf{200} & 31.57 & \textbf{200} & 10.82 & \textbf{200} & 0.97 & \textbf{200} & 0.91 & \textbf{200} & \textbf{0.16} \\
18 & 500 & 12311 & \textbf{500} & 819.77 & \textbf{500} & 61.33 & \textbf{500} & 11.46 & \textbf{500} & 4.46 & \textbf{500} & 3.9 & \textbf{500} & \textbf{1.63} \\
18 & 1000 & 24643 & \textbf{1000} & 901.1 & \textbf{1000} & 61.3 & 999 & 12.01 & \textbf{1000} & 20.09 & 976 & 13.8 & 979 & \textbf{9.39} \\
18 & 2000 & 49572 & 1952 & 901.78 & 1947 & 61.28 & 1940 & \textbf{17.53} & \textbf{1957} & 940.1 & 1822 & 39.51 & 1826 & 50.26 \\
18 & 5000 & 125633 & \textbf{3794} & 912.57 & 3742 & 66.58 & 3740 & \textbf{61.78} & 3427 & 1109.11 & 3427 & 208.89 & 3454 & 498.95 \\
18 & 7000 & 177779 & \textbf{4313} & 925.96 & 4302 & 172.37 & 4269 & \textbf{147.96} & 3897 & 1426.06 & 3897 & 531.45 & 3006 & 913.83 \\
18 & 10000 & 256733 & \textbf{4739} & 1077.63 & 4671 & 446.56 & 4540 & \textbf{346.1} & 4362 & 3492.77 & 4362 & 2551.72 & 790 & 921.53 \\
\bottomrule
\end{tabular}
}
\end{center}
\label{tab:skysat_scheduling_results}
\end{table}

\Cref{tab:walker_scheduling_results} presents the scheduling outcomes for each of the walker constellations. The number of scheduled collects $|X^S|$ returned by each method is listed for each scenario along with the solver runtime $t_{solve}$. The runtime is measured as time from execution start to when the solution call returns. The best solution, in terms of number of scheduled collects $\mathcal{O} = |X^S|$, is highlighted in bold. The solver with the lowest runtime is also highlighted. Because the repeat collection constraint prevents duplicate scheduling of any single request, there is an upper bound on any scenario, which is $|X^S| \leq |R|$. Therefore, any solution with $|X^S| = |R|$ is an optimal schedule. We cannot say if the the solution is unique, though it is in general not expected to be because there can be multiple maximal independent sets as seen in \Cref{fig:feasibility_contradictions}(d).

\begin{figure}[htb]
\begin{center}

\begin{subfigure}{.5\textwidth}
\begin{center}
\includegraphics[width=\textwidth]{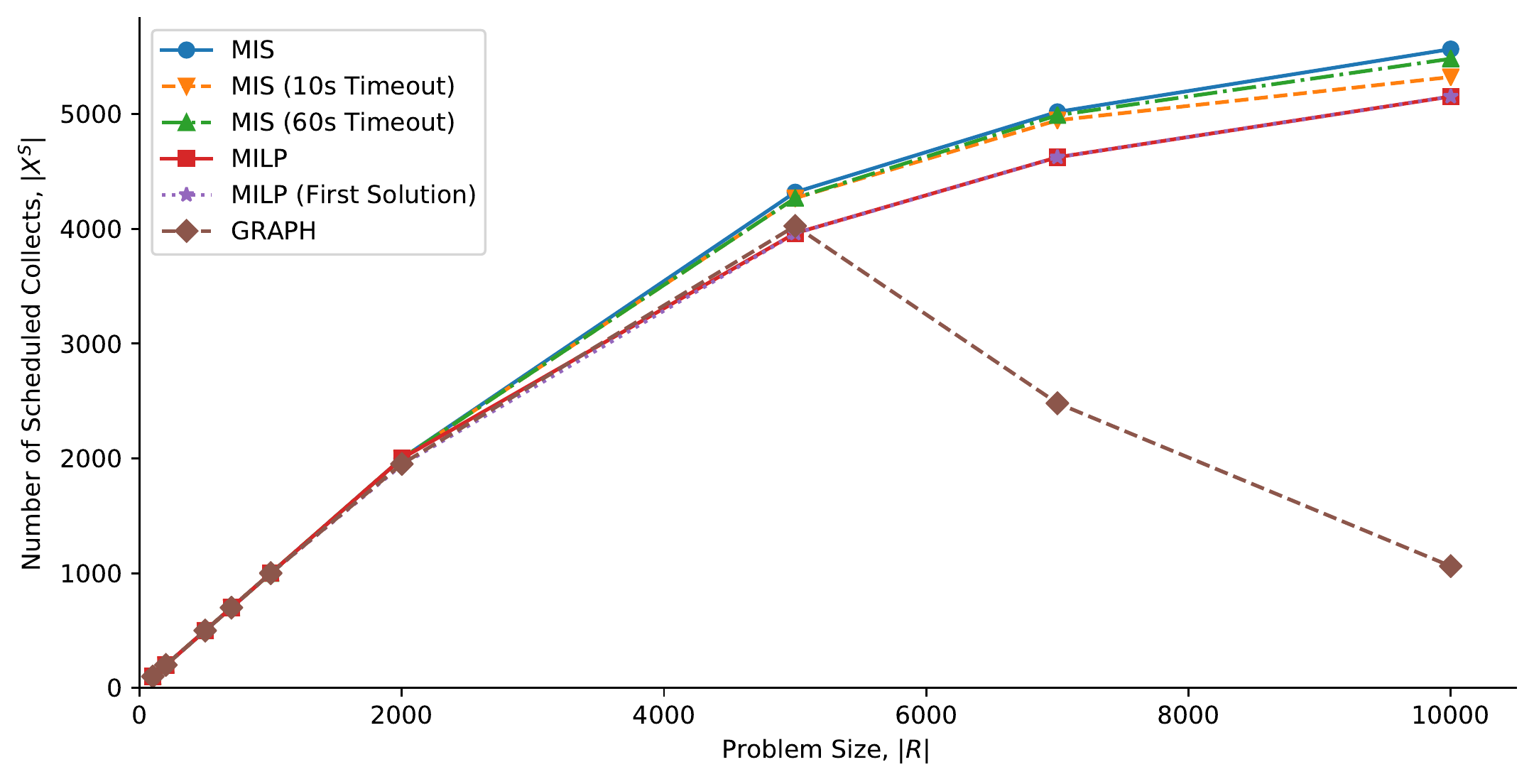}	
\end{center}
\end{subfigure}%
\begin{subfigure}{.5\textwidth}
\begin{center}
\includegraphics[width=\textwidth]{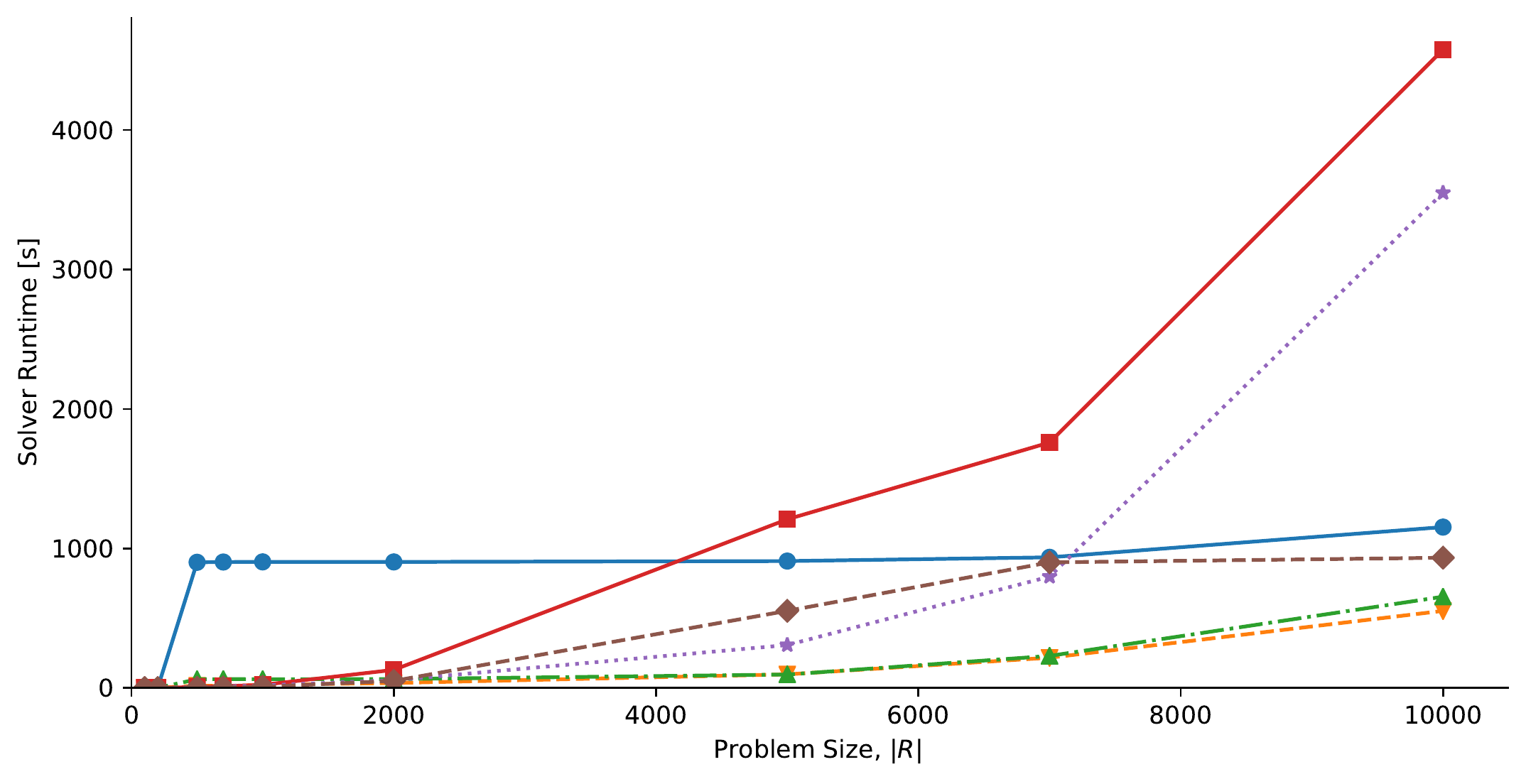}	
\end{center}
\end{subfigure}

\end{center}
\caption{\label{fig:sats24_scaling}Comparison of scheduling techniques for largest, 24 satellite, walker constellation. The number of scheduled collects (left) and solver runtime (right) are presented.}
\end{figure}

There are a few interesting trends in the table. First, for small problem sizes, those with $500$ or fewer requests, all scheduling algorithms aside from the MILP ($1\text{st}$ solution) and graph traversal algorithms always find the optimal solution. The sole exception to this trend occurs in the 4-satellite, 500-request test case where the MIS solvers are unable to find a schedule of $|X^S| = 385$. Instead, the MIS solvers find $|X^S| = 384$. The fact that MIS solutions do not find the best solution in all cases is expected since the MIS algorithm relies on local search, evolutionary, and recursive techniques that do not provide optimality guarantees. However, the MIS approaches find the best schedule in 33 of the 36 test cases where $|R| \leq 500$. The table also shows that, for small problems, the MIS approaches takes longer to arrive at a solution than the MILP or graph approaches. For small problems, the MIS solution takes tens of seconds to solve compared to second or sub-second solutions of the MILP and graph solvers. The longer solution times are attributed to algorithm spending time recursing on the final solution in an attempt to find a larger maximal independent set. This behavior is observed in other applications of the algorithm where an initial large independent set can be found quickly, but finding a maximal independent set takes longer \cite{lamm2017finding}. This indicates that the MIS approach to satellite scheduling could be improved by providing the solver an informed upper optimality bound so that the algorithm can exit early once a maximal solution is found. Unlike most maximum independent set problems that do not have known upper bounds, knowledge of the satellite scheduling problem allows for the development of bounds. This could prevent the algorithm from spending long periods of time attempting to find a better solution that does not exist due to the presence of the repeat constraint.

\Cref{fig:sats24_scaling} shows the performance trends observed in the table for the largest, 24 satellite, scenario. As $|R|$ increases, all three MIS methods are observed to schedule equal or greater numbers of collects than either the MILP or graph traversal methods. Additionally, the difference in $|X^S|$ between the MIS and MILP solutions is seen to increase with the number of requests. The MIS methods are consistently observed to have lower runtimes than other methods. The figure shows that for $|R| > 500$ the MIS solver uses up to the allowed maximum, 900 second, solve duration. However, there is no a significant increase in the number of scheduled collects found by the default MIS solver compared to the 10 second and 60 second time-limited MIS solvers, meaning that by setting the allowed solve time lower, it is possible to significantly reduce the runtime at the cost of a small loss of solution quality.

\Cref{tab:walker_scheduling_results} also shows that the graph traversal solution is frequently unable to find optimal schedules. This can be attributed to the deficiencies of feasibility-based approaches in coordinating actions of multiple agents discussed in \Cref{sec:graph_interpretation}. Graph traversal solves each satellite in order so satellite's schedule affects the ability of subsequent satellites to achieve optimality due to inability to coordinate activities across spacecraft and time horizon. The inability of the graph traversal algorithm to find maximal schedules for small numbers of requests and spacecraft (e.g. $|S| = 4, |R| = 200$) supports the hypothesis that the infeasibility interpretation is better suited for constellation scheduling as it can simultaneously coordinate across spacecraft and the entire time horizon.

For large problem sizes, those with $|R| \geq 5000$, the advantage of the maximum independent set approach becomes clear. In all cases, the MIS solver schedules more collects (larger $|X^S|$) than either of the MILP approaches. In fact, all MIS approaches, even the 10-second timeout configuration, find solutions with larger $|X^S|$ than either of the MILP approaches and in less time. For the largest test case, the 24-satellite, 10,000-request scenario, the MIS algorithm schedules 5,566 collects in \SI{1153.33}{\second} while the MILP solution schedules 5,153 collects in \SI{4576.03}{\second}. The MIS solver improves scheduled collections by 8.01\% and in 74.49\% less time. Both the MILP and MILP ($1\text{st}$ solution) solvers return the same solution size of 5,153 for this test case. This is because both MILP approaches spend the majority of time pre-solving the problem and there is not enough time before the termination signal for the MILP solver to complete a second iteration of the simplex solver beyond the first. 

\Cref{tab:skysat_scheduling_results} presents the results of scheduling requests for the current Skysat constellation of imaging spacecraft, which is a mixed inclination constellation of both sun-synchronous and mid-inclination orbits. The trends observed for the Walker constellation tests also hold for the Skysat constellation. The MIS solution is able to find the optimal scheduling solution for scenarios with $|R| \leq 1000$. For scenarios with $|R| \geq 5000$, all three maximum independent set solvers schedule return larger $|X^S|$ than either of the MILP solutions. For the largest 10,000-request test case the MIS approach schedules 4,739 collects in \SI{1077.63}{\second} while the MILP solution schedules 4,362 collects in \SI{3492.77}{\second}, an 8.64\% improvement in number of collects and a 69.14\% reduction in computation time. When compared to the MILP ($1\text{st}$ solution) solver, it is a 8.64\% improvement in number of collects and a 57.76\% improvement in runtime, slightly smaller but still a significant improvement. The MIS solver time limit becomes hyper parameter satellite operators can use to tune the algorithm and trade between number of scheduled collects and runtime. Comparing the MIS (\SI{10}{\second}) solution to the regular MIS solution, the former scheduled 4.19\% fewer collects, but only takes \SI{346.1}{\second}, the least overall runtime of any solution method tested. Compared to the MILP ($1\text{st}$ solution), the MIS (\SI{10}{\second}) solution returns a 4.08\% improved schedule but in 86.43\% less time. Which is a slight improvement in overall solution quality and significant improvement in solution speed.


\section{Conclusion}
\label{sec:conclusion}

This paper poses satellite task scheduling as a maximal independent set problem. The difficulty of the scheduling problem depends on the number of collection opportunities over the planning horizon. The number of collections in the scheduling horizon grows with the constellation size, making the scheduling problem more challenging for large constellations. Task scheduling problem can be represented as a sparse, undirected graph that allows for efficient deconfliction and coordination of activities between multiple satellites. The problem is solved using the ReduMIS algorithm, which efficiently finds large independent sets that correspond to satellite task plans.

Simulation results show that the maximum independent set method finds optimal or near-optimal solutions all problem sizes. For problems with 500 requests or fewer, the MIS methods finds optimal solutions in comparable, though slightly longer, time than the baseline mixed-integer linear program and graph-traversal methods. For larger problem sizes, the maximum independent set method outperforms the baseline methods in both number of scheduled collects and execution time. Additionally, by imposing a time limit on the solution time it is possible to trade between solution quality and run time. In the largest test scenario, which considers up to 24 satellites, 10,000 requests, and 359,170 collection opportunities, the maximum independent set method improves the number of scheduled collects by 8\% while simultaneously reducing the scheduling time by 75\% compared to the MILP method. The formulation is applied to the current Skybox constellation and delivers a 4.08\% improved schedule but in 86.43\% less time than the fastest MILP method.

Areas for future work include informing the solution method with domain-derived upper bounds to further reduce solve time, including satellite power and data resource constraints as part of the problem formulation, and extending the solution to weighted independent sets to allow non-uniform request priorities.

%

\section*{Acknowledgments}
The authors would like to thank Zouhair Mahboubi for many interesting discussions on the satellite scheduling problem. We would also like to thank Christian Lenz for his continued support and encouragement in pursing this research. Finally, we would like to thank the United States Geospatial Intelligence Foundation for both financial support and providing a welcoming community that enthusiastically encouraged this research.

\bibliography{deddy_research}

\begin{thebibliography}{30}
\newcommand{\enquote}[1]{``#1''}
\providecommand{\natexlab}[1]{#1}
\providecommand{\url}[1]{\texttt{#1}}
\providecommand{\urlprefix}{URL }
\expandafter\ifx\csname urlstyle\endcsname\relax
  \providecommand{\doi}[1]{doi:\discretionary{}{}{}#1}\else
  \providecommand{\doi}{doi:\discretionary{}{}{}\begingroup
  \urlstyle{rm}\Url}\fi

\bibitem[{Henely et~al.(2019)Henely, Baldwin-Pulcini, and
  Smith}]{henely2019turning}
Henely, S., Baldwin-Pulcini, B., and Smith, K., \enquote{Turning off the
  lights: Automating SkySat mission operations,} \emph{Small Satellite
  Conference}, 2019.

\bibitem[{Boshuizen et~al.(2014)Boshuizen, Mason, Klupar, and
  Spanhake}]{boshuizen2014results}
Boshuizen, C., Mason, J., Klupar, P., and Spanhake, S., \enquote{Results from
  the planet labs flock constellation,} \emph{Small Satellite Conference},
  2014.

\bibitem[{Irisov et~al.(2018)Irisov, Nguyen, Duly, Masters, Nogues-Correig,
  Tan, Yuasa, and Ector}]{irisov2018recent}
Irisov, V., Nguyen, V., Duly, T., Masters, D.~S., Nogues-Correig, O., Tan, L.,
  Yuasa, T., and Ector, D.~R., \enquote{Recent Ionosphere collection results
  from Spire's 3U CubeSat GNSS-RO constellation,} \emph{American Geophysical
  Union Fall Meeting}, 2018.

\bibitem[{{National Research Council Space Studies
  Board}(2007)}]{board2007earth}
{National Research Council Space Studies Board}, \emph{Earth science and
  applications from space: National imperatives for the next decade and
  beyond}, National Academies Press, 2007.

\bibitem[{Popkin(2017)}]{popkin2017commercial}
Popkin, G., \enquote{Commercial space sensors go high-tech,} \emph{Nature},
  Vol. 545, No. 7655, 2017, pp. 397--398.

\bibitem[{Stringham et~al.(2019)Stringham, Farquharson, Castelletti, Quist,
  Riggi, Eddy, and Soenen}]{stringham2019capella}
Stringham, C., Farquharson, G., Castelletti, D., Quist, E., Riggi, L., Eddy,
  D., and Soenen, S., \enquote{The Capella X-band SAR constellation for rapid
  imaging,} \emph{IEEE International Geoscience and Remote Sensing Symposium},
  2019.

\bibitem[{Sarda et~al.(2018)Sarda, CaJacob, Orr, and Zee}]{sarda2018making}
Sarda, K., CaJacob, D., Orr, N., and Zee, R., \enquote{Making the invisible
  visible: Precision RF-emitter geolocation from space by the hawkeye 360
  pathfinder mission,} \emph{Small Satellite Conference}, 2018.

\bibitem[{Brown and Eremenko(2006)}]{brown2006value}
Brown, O., and Eremenko, P., \enquote{The value proposition for fractionated
  space architectures,} \emph{AIAA Space Forum}, 2006.

\bibitem[{Wertz et~al.(2011)Wertz, Everett, and Puschell}]{wertz2011space}
Wertz, J.~R., Everett, D.~F., and Puschell, J.~J., \emph{Space Mission
  Engineering: The New SMAD}, Microcosm Press, 2011.

\bibitem[{Garey and Johnson(1979)}]{garey1979computers}
Garey, M.~R., and Johnson, D.~S., \emph{Computers and Intractability: A Guide
  to the Theory of NP-Completeness}, W.H Freeman and Company, 1979.

\bibitem[{Hall and Magazine(1994)}]{hall1994maximizing}
Hall, N.~G., and Magazine, M.~J., \enquote{Maximizing the Value of a Space
  Mission,} \emph{European Journal of Operational Research}, Vol.~78, No.~2,
  1994, pp. 224--241.

\bibitem[{Lema{\^i}tre et~al.(2002)Lema{\^i}tre, Verfaillie, Jouhaud, Lachiver,
  and Bataille}]{lemaitre2002selecting}
Lema{\^i}tre, M., Verfaillie, G., Jouhaud, F., Lachiver, J.-M., and Bataille,
  N., \enquote{Selecting and Scheduling Observations of Agile Satellites,}
  \emph{Aerospace Science and Technology}, Vol.~6, No.~5, 2002, pp. 367--381.

\bibitem[{Bianchessi(2005)}]{bianchessi2005earth}
Bianchessi, N., \enquote{Earth Observation Satellites: Models and Algorithms,}
  Ph.D. thesis, Universit{\`a} Delgi Studi di Milano, 2005.

\bibitem[{Beaumet et~al.(2011)Beaumet, Verfaillie, and
  Charmeau}]{beaumet2011feasibility}
Beaumet, G., Verfaillie, G., and Charmeau, M.-C., \enquote{Feasibility of
  Autonomous Decision Making on Board an Agile Earth-Observing Satellite,}
  \emph{Computational Intelligence}, Vol.~27, No.~1, 2011, pp. 123--139.

\bibitem[{Lamm et~al.(2017)Lamm, Sanders, Schulz, Strash, and
  Werneck}]{lamm2017finding}
Lamm, S., Sanders, P., Schulz, C., Strash, D., and Werneck, R.~F.,
  \enquote{Finding near-optimal independent sets at scale,} \emph{Journal of
  Heuristics}, Vol.~23, No.~4, 2017, pp. 207--229.

\bibitem[{Harrison et~al.(1999)Harrison, Price, and
  Philpott}]{harrison1999task}
Harrison, S., Price, M., and Philpott, M., \enquote{Task scheduling for
  satellite based imagery,} \emph{Workshop of the UK Planning and Scheduling
  Special Interest Group}, University of Sanford, 1999.

\bibitem[{Martin(2002)}]{martin2002satellite}
Martin, W., \enquote{Satellite Image Collection Optimization,} \emph{Optical
  Engineering}, Vol.~41, 2002, pp. 2083--2087.

\bibitem[{Iacopino(2012)}]{iacopino2012highly}
Iacopino, C., \enquote{Highly responsive MPS for dynamic EO scenarios,}
  \emph{AIAA SpaceOps Conference}, 2012.

\bibitem[{Bianchessi and Righini(2006)}]{bianchessi2006mathematical}
Bianchessi, N., and Righini, G., \enquote{A mathematical programming algorithm
  for planning and scheduling an earth observing SAR constellation,}
  \emph{International Workshop on Planning and Scheduling for Space (IWPSS)},
  2006.

\bibitem[{Bianchessi et~al.(2007)Bianchessi, Cordeau, Desrosiers, Laporte, and
  Raymond}]{bianchessi2007heuristic}
Bianchessi, N., Cordeau, J.-F., Desrosiers, J., Laporte, G., and Raymond, V.,
  \enquote{A heuristic for the multi-satellite, multi-orbit and multi-user
  management of Earth observation satellites,} \emph{European Journal of
  Operational Research}, Vol. 177, No.~2, 2007, pp. 750--762.

\bibitem[{Iacopino et~al.(2013{\natexlab{a}})Iacopino, Palmer, Brewer,
  Policella, and Donati}]{iacopino2013eo}
Iacopino, C., Palmer, P., Brewer, A., Policella, N., and Donati, A.,
  \enquote{EO Constellation MPS based on ant colony optimization algorithms,}
  \emph{IEEE International Conference on Recent Advances in Space
  Technologies}, 2013{\natexlab{a}}.

\bibitem[{Iacopino et~al.(2013{\natexlab{b}})Iacopino, Palmer, Policella,
  Donati, and Brewer}]{iacopino2013self}
Iacopino, C., Palmer, P., Policella, N., Donati, A., and Brewer, A.,
  \enquote{Self-organizing MPS for dynamic EO constellation scenarios,}
  \emph{International Workshop on Planning and Scheduling for Space (IWPSS)},
  2013{\natexlab{b}}.

\bibitem[{Augenstein(2014)}]{augenstein2014optimal}
Augenstein, S., \enquote{Optimal Scheduling of Earth-Imaging Satellites with
  Human Collaboration via Directed Acyclic Graphs,} \emph{AAAI Spring Symposium
  on the Intersection of Robust Intelligence and Trust in Autonomous Systems},
  2014.

\bibitem[{Augenstein et~al.(2016)Augenstein, Estanislao, Guere, and
  Blaes}]{augenstein2016optimal}
Augenstein, S., Estanislao, A., Guere, E., and Blaes, S., \enquote{Optimal
  Scheduling of a Constellation of Earth-Imaging Satellites for Maximal Data
  Throughput and Efficient Human Management,} \emph{International Conference on
  Automated Planning and Scheduling (ICAPS)}, 2016.

\bibitem[{Nag et~al.(2018)Nag, Li, and Merrick}]{nag2018scheduling}
Nag, S., Li, A.~S., and Merrick, J.~H., \enquote{Scheduling Algorithms for
  Rapid Imaging Using Agile Cubesat Constellations,} \emph{Advances in Space
  Research}, Vol.~61, No.~3, 2018, pp. 891--913.

\bibitem[{Van~Etten et~al.(2020)Van~Etten, Shermeyer, Hogan, Weir, and
  Lewis}]{van2020road}
Van~Etten, A., Shermeyer, J., Hogan, D., Weir, N., and Lewis, R., \enquote{Road
  network and travel time extraction from multiple look angles with SpaceNet
  data,} \emph{arXiv preprint arXiv:2001.05923}, 2020.

\bibitem[{Vallado et~al.(2006)Vallado, Crawford, Hujsak, and
  Kelso}]{vallado2006revisiting}
Vallado, D., Crawford, P., Hujsak, R., and Kelso, T., \enquote{Revisiting
  Spacetrack Report \#3,} \emph{AIAA/AAS Astrodynamics Specialist Conference
  and Exhibit}, 2006.

\bibitem[{Eddy and Kochenderfer(2020)}]{eddyMarkov2020}
Eddy, D., and Kochenderfer, M., \enquote{Markov Decision Processes for
  multi-objective satellite task planning,} \emph{IEEE Aerospace Conference},
  2020.

\bibitem[{Andrade et~al.(2012)Andrade, Resende, and Werneck}]{andrade2012fast}
Andrade, D.~V., Resende, M.~G., and Werneck, R.~F., \enquote{Fast local search
  for the maximum independent set problem,} \emph{Journal of Heuristics},
  Vol.~18, No.~4, 2012, pp. 525--547.

\bibitem[{Lamm et~al.(2015)Lamm, Sanders, and Schulz}]{lamm2015graph}
Lamm, S., Sanders, P., and Schulz, C., \enquote{Graph partitioning for
  independent sets,} \emph{International Symposium on Experimental Algorithms},
  2015.

\end{thebibliography}

\end{document}